\documentclass[journal,12pt,onecolumn]{IEEEtran}
\usepackage{amsmath,amsfonts}
\usepackage{algorithmic}
\usepackage{algorithm}
\usepackage{array}
\usepackage{textcomp}
\usepackage{stfloats}
\usepackage{url}
\usepackage{verbatim}
\usepackage{graphicx}
\usepackage{cite}
\usepackage{subcaption}
\usepackage{mathtools}
\usepackage{xcolor}
\usepackage{placeins}
\usepackage{bbm}
\usepackage[percent]{overpic}
\usepackage{transparent}

\usepackage{booktabs} % for professional tables
\usepackage{amssymb}
\usepackage{enumitem}
\usepackage{mathtools}
\usepackage{amsthm}

\theoremstyle{plain}
\newtheorem{theorem}{Theorem}[section]

\theoremstyle{definition}

\theoremstyle{remark}

\begin{document}

\title{TurbuGAN: An Adversarial Learning Approach to Spatially-Varying Multiframe Blind Deconvolution with Applications to Imaging Through Turbulence}

\author{Brandon Y. Feng\IEEEauthorrefmark{1}, Mingyang Xie\IEEEauthorrefmark{1}, Christopher A. Metzler
%\author{IEEE Publication Technology,~\IEEEmembership{Staff,~IEEE,}
        % <-this % stops a space
%\thanks{This paper was produced by the IEEE Publication Technology Group. They are in Piscataway, NJ.}% <-this % stops a space
\thanks{*Equal Contributions. All authors are with the Department of Computer Science, University of Maryland, College Park, MD, 20742, USA (e-mail: \{yfeng97, mingyang, metzler\}@umd.edu). This work was supported in part by the AFOSR Young Investigator Program Award FA9550-22-1-0208.}
%\thanks{Manuscript received April 19, 2021; revised August 16, 2021.}
}

% The paper headers
%\markboth{Journal of \LaTeX\ Class Files,~Vol.~14, No.~8, August~2021}%
%{Shell \MakeLowercase{\textit{et al.}}: A Sample Article Using IEEEtran.cls for IEEE Journals}

%\IEEEpubid{0000--0000/00\$00.00~\copyright~2022 IEEE}
% Remember, if you use this you must call \IEEEpubidadjcol in the second
% column for its text to clear the IEEEpubid mark.

\maketitle

\begin{abstract}
We present a self-supervised and self-calibrating multi-shot approach to imaging through atmospheric turbulence, called TurbuGAN. 
Our approach requires no paired training data, adapts itself to the distribution of the turbulence, leverages domain-specific data priors, and can generalize from tens to thousands of measurements. 
We achieve such functionality through an adversarial sensing framework adapted from CryoGAN~\cite{Gupta2021CryoGAN}, which uses a discriminator network to match the distributions of captured and simulated measurements. 
Our framework builds on CryoGAN by (1) generalizing the forward measurement model to incorporate physically accurate and computationally efficient models for light propagation through anisoplanatic turbulence, (2) enabling adaptation to slightly misspecified forward models, and (3) leveraging domain-specific prior knowledge using pretrained generative networks, when available. 
We validate TurbuGAN on both computationally simulated and experimentally captured images distorted with anisoplanatic turbulence.
\end{abstract}
\begin{figure*}[!ht]
    \centering
    \includegraphics[width=0.95\textwidth]{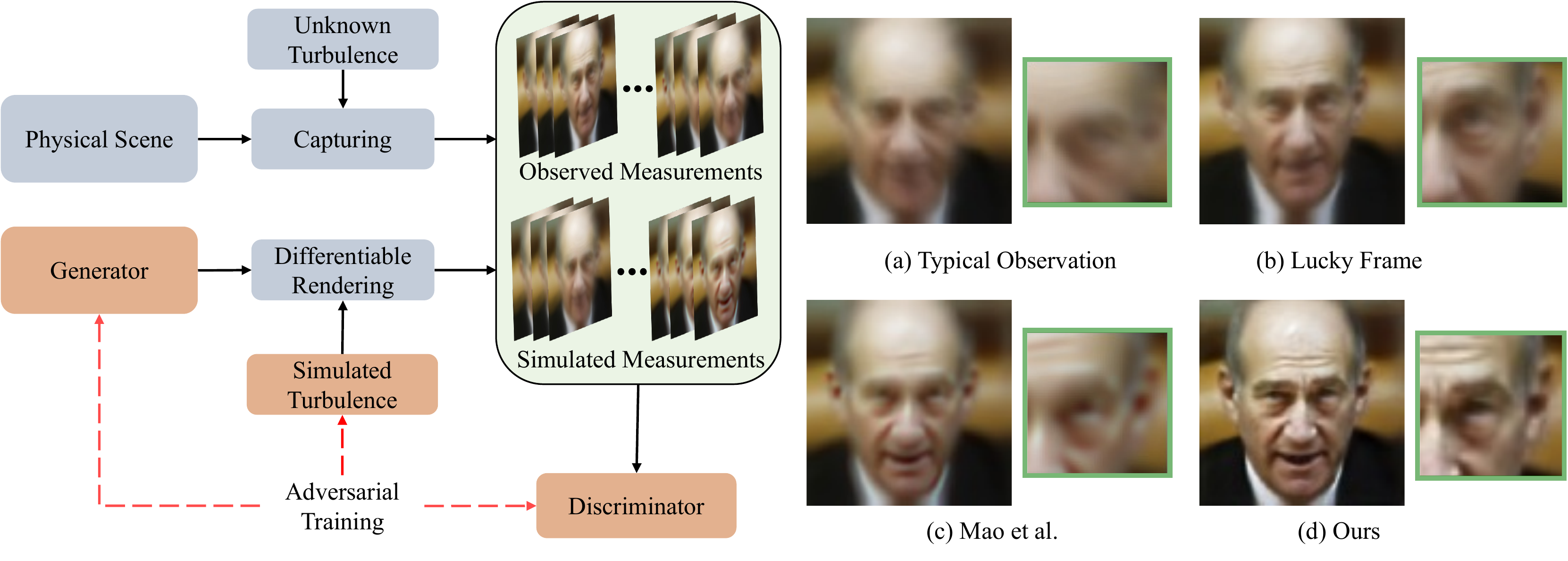}
    \caption{{\bf TurbuGAN Framework and Reconstructions.}
    Capturing images through turbulence often leads to highly distorted and blurry observations (a).
    While some individual ``lucky'' frames (b) may be sharper than others, one still needs to combine information from many frames to accurately reconstruct the scene. %prior methods often try to select a sharp lucky frame (b), it still fails to preserve many fine details from the original scene.
    TurbuGAN (left) uses a collection of distorted measurements and produce reconstructions (d) that are sharper and contain fewer artifacts than the state-of-the-art method (c).
    TurbuGAN achieves this using adversarial sensing: A generator serves as the neural representation of the distortion-free scene, whose output is fed into a physically accurate and differentiable rendering engine that models the effects of atmospheric turbulence to produce simulated measurements. 
    A discriminator compares the distribution of these measurements with the distribution of true observations of the scene. The discriminator's feedback is used to update the generator (and optionally the turbulence simulator). The scene is finally reconstructed when the simulated measurements become indistinguishable from the true observations. \vspace{-15pt}
    } \label{fig:teaser}
\end{figure*}

\begin{IEEEkeywords}
Adversarial training, multiframe blind deconvolution, atmospheric turbulence.
\end{IEEEkeywords}

\newcommand{\etal}{{\em et al.}}
\newcommand{\eg}{{\em e.g.}}
\newcommand{\degree}{\ensuremath{^{\circ}} }

\newcommand{\SevenColHspace}{67pt}
\newcommand{\SevenColSubfigWidth}{67pt}
\newcommand{\SevenColSubfig}[1]{	\begin{subfigure}{\SevenColSubfigWidth}
	\centering
	\includegraphics[width=\SevenColSubfigWidth, height=\SevenColSubfigWidth]{#1}
	\vspace{1pt}
\end{subfigure}
}
\newenvironment{paracolor}{\par\color{blue}}{\par}

\newcommand{\SevenColSubfigwCaption}[2]{
\begin{subfigure}{\SevenColSubfigWidth}
	\centering
	\includegraphics[width=\SevenColSubfigWidth, height=\SevenColSubfigWidth]{#1}
	\caption{#2}
\end{subfigure}
}

\newcommand{\edit}[1]{\textcolor{black}{#1}}
\newcommand{\reviewer}[1]{\textcolor{\definecolor{darkblue}{rgb}{0,0,0.8}}{#1}}
 
 \definecolor{darkblue}{rgb}{0,0,0.5}
\newcommand{\reviewercomments}[1]{\noindent\textcolor{darkblue}{#1}}

\newcommand{\brandon}[1]{\textcolor{brown}{#1}}

\newcommand{\ThreeColSubfigWidth}{130pt}
\newcommand{\ThreeColSubfig}[1]{	
\begin{subfigure}{\ThreeColSubfigWidth}
	\centering
	\includegraphics[width=\ThreeColSubfigWidth, height=\ThreeColSubfigWidth]{#1}
\end{subfigure}
}
\newcommand{\ThreeColSubfigwCaption}[2]{	
\begin{subfigure}{\ThreeColSubfigWidth}
	\centering
	\includegraphics[width=\ThreeColSubfigWidth, height=\ThreeColSubfigWidth]{#1}
	\caption{#2}
\end{subfigure}
}

\newcommand{\ThreeColSubfigwBox}[2]{
\begin{subfigure}{\ThreeColSubfigWidth}
	\centering
	\includegraphics[width=\ThreeColSubfigWidth, height=\ThreeColSubfigWidth]{#1}
    $\underbracket[0pt][1mm]{\hspace{0.3\linewidth}}_%
    {\substack{\vspace{-2cm}\\ {\textcolor{white}{\hspace{3.5cm}\text{#2}}}}}$
    \vspace{-0.9cm}
\end{subfigure}
}

\newcommand{\FourColSubfigWidth}{100pt}
\newcommand{\FourColSubfig}[1]{	
\begin{subfigure}{\FourColSubfigWidth}
	\centering
	\includegraphics[width=\FourColSubfigWidth, height=\FourColSubfigWidth]{#1}
\end{subfigure}
}

\newcommand{\FiveColHspace}{92pt}
\newcommand{\FiveColSubfigWidth}{92pt}
\newcommand{\FiveColSubfig}[1]{	\begin{subfigure}{\FiveColSubfigWidth}
	\centering
	\includegraphics[width=\FiveColSubfigWidth, height=\FiveColSubfigWidth]{#1}
	\vspace{1pt}
\end{subfigure}
}

\newcommand{\FiveColSubfigwCaption}[2]{	
\begin{subfigure}{\FiveColSubfigWidth}
	\centering
	\includegraphics[width=\FiveColSubfigWidth, height=\FiveColSubfigWidth]{#1}
	\caption{#2}
\end{subfigure}
}

\newcommand{\SixColSubfigWidth}{91pt}
\newcommand{\SixColSubfigwCaption}[2]{
\begin{subfigure}{\SixColSubfigWidth}
	\centering
	\includegraphics[width=\SixColSubfigWidth, height=\SixColSubfigWidth]{#1}
	\caption{#2}
\end{subfigure}
}
\newtheorem{custom_corollary}{Corollary}[theorem]

\newcommand{\SixxColSubfigWidth}{70pt}
\newcommand{\SixColSubfig}[1]{	
\begin{subfigure}{\SixxColSubfigWidth}
	\centering
	\includegraphics[width=\SixxColSubfigWidth, height=\SixxColSubfigWidth]{#1}
\end{subfigure}
}

\section{Introduction}
\label{Introduction}
Atmospheric turbulence causes long-range imaging systems to capture blurry and distorted measurements.
These distortions are caused by the heterogeneous refractive index of the atmosphere: When light propagates through the atmosphere, this heterogeneity will induce spatially-varying phase delays in the wavefront, which in turn determine the optical system's point spread function (PSF). 
When the heterogeneity is distributed over a volume, the turbulence is said to be anisoplanatic and the PSF is spatially-varying.
Accordingly, imaging through anisoplanatic turbulence can be described by the equation
\edit{
\begin{align}
% \pmb{y}_i(u,v) = \pmb{h}_i(u,v) * \pmb{x} + \pmb{\epsilon}_i,
\pmb{y}_i[u,v] = (\pmb{h}_i(u,v) * \pmb{x})[u,v] + \pmb{\epsilon}_i[u,v],
\label{eq:1}
\end{align}
where $u$ and $v$ are pixel coordinates; $\pmb{y}_i \in \mathbb{R}^{{N}\times{N}}$ is an observed blurry image; $\pmb{h}_i \colon (u,v)\in\mathbb{R}^{N\times N} \mapsto \pmb{h}_i(u,v)\in\mathbb{R}^{M\times M}$ is a bivariate matrix-valued function which defines the spatially-varying blur kernel, which maps pixel coordinates $(u,v)$ to the blur kernel  $\pmb{h}_i(u,v)$ at that location;} $\pmb{x} \in \mathbb{R}^{{N}\times{N}}$ is the target being imaged; $\pmb{\epsilon}_i \in \mathbb{R}^{{N}\times{N}}$ is (possibly signal dependent) noise; $N$ is the image size in pixels; and $M$ is the kernel size in pixels. 
When imaging a stationary target through dynamic turbulence, the signal $\pmb{x}$ is fixed while at each time step the measurements $\pmb{y}_1,\pmb{y}_2,\dots~\pmb{y}_L$ experience different blur kernels $\pmb{h}_1,\pmb{h}_2,\dots~\pmb{h}_L$.

The central challenge posed by the forward model described by Equation~\eqref{eq:1} is that the underlying turbulence is {\em stochastic} across different locations and instances of time.
Without prior information or measurements such as a guidestar~\cite{beckers1993adaptive}, the spatially-varying blur kernels $\{\pmb{h}_i\}_{i=1}^L$ are unknown, and recovering $\pmb{x}$ requires solving a spatially-varying multiframe blind deconvolution problem.

To tackle this problem, classical algorithms often explicitly estimate or marginalize over the latent variables (blur kernels) which characterize the turbulence associated with each measurement.
Such strategies are computationally intensive and impractical when the turbulence is spatially-varying and/or the number of measurements becomes large. 
Alternatively, one can train an application-specific deep neural network to reconstruct sharp images from distorted measurements.
However, such networks rely on large quantities of representative training data and cannot be easily adapted to new turbulence models at test time.

% In this work, we develop a self-supervised reconstruction algorithm that can be adapted at test time to the particular turbulence distribution, without retraining. Additionally we seek a method that can leverage pre-trained application-agnostic generative models, like StyleGAN, so as to take advantage of strong learning-based priors.
% require a large set of training data with labelled ground truth, which are scarce for atmospheric turbulence.
% Moreover, deep learning methods generally perform poorly when applied to out-of-distribution test data, such as turbulence strengths unseen in the training set. 
%\textcolor{red}{Should mention how other methods work and why they are inadequate}

\subsection{Contributions}
We present TurbuGAN, a new self-supervised spatially-varying multiframe blind deconvolution algorithm for imaging through atmospheric turbulence.
The proposed algorithm uses an adversarial sensing framework shown in Fig.~\ref{fig:teaser}, in which a neural representation of a scene and a physically-accurate differentiable forward model are combined to generate synthetic measurements. 
A discriminator network is then used to compare the {\em distributions} of the synthesized and experimental measurements to provide a feedback signal that improves the quality of synthesized measurements. 
Adversarial sensing, which was first developed for unknown-view-tomography in the context of single-particle cryogenic electron microscopy (cryo-EM)~\cite{Gupta2021CryoGAN}, is explained in detail in Section~\ref{Methods}.

To our knowledge, this is the first application of adversarial sensing to multiframe blind deconvolution or imaging through atmospheric turbulence. 
By leveraging adversarial sensing, TurbuGAN provides several distinct advantages over previous methods. 
Unlike classical algorithms, TurbuGAN requires neither reconstructing nor marginalizing over the latent variables which characterize the turbulence associated with each measurement.
Unlike supervised deep learning methods, TurbuGAN does not rely on a large corpus of paired training data and is thus robust to out-of-distribution test data or turbulence strengths. 
Moreover, while many (but not all, \eg~\cite{aittala2018burst}) supervised deep learning methods make predictions based on a fixed number of input frames in a single-shot fashion, TurbuGAN is able to iteratively refine its prediction given an arbitrary number of measurements.

Our largest contributions can be summarized as follows.
\begin{enumerate}[topsep=0pt,itemsep=-1ex,partopsep=1ex,parsep=1ex]
    \item We present a novel adversarial sensing framework for the problem of multiframe imaging through anisoplanatic turbulence.
    \item We demonstrate adversarial sensing can be combined with off-the-shelf generative networks (\eg~StyleGAN~\cite{karras2020analyzing} and StyleGAN-XL~\cite{Sauer2022StyleGANXLSS}) to impose strong but specialized priors on the reconstructed images.
    \item We demonstrate adversarial sensing can be combined with untrained generative networks (\eg~DIP~\cite{Ulyanov2018DeepIP}) to impose weak but flexible priors on the reconstructed images.
    \item We extend adversarial sensing to deal with misspecified forward model and show that this extension allows us to adapt to unknown turbulence conditions.
    \item We validate our method on both computationally simulated and experimentally captured data.% and demonstrate exceptional performance.
\end{enumerate}

\subsection{Limitations}
% \textcolor{red}{This needs a rewrite.}
% Our method updates the estimated underlying image so that its synthetically-blurred results are closer to the distributions of the actual observations. 
% As such, the method's performance would likely be unremarkable when only a small set of observations is available and thus the problem is highly under-determined.
While our work represents a fundamentally new approach to imaging through turbulence,  it does come with a few notable limitations. 
First, our current work is designed to work with multiple measurements of a { static} object. \edit{Though our method can be extended to reconstruct simple dynamic scenes such as binary moving digits, it fails to extend to real-world dynamic scenes of complex objects. } % where the underlying $\pmb{x}$ changes between measurements $\pmb{y}$. 
%Second, while the forward model used in this paper has been experimentally validated~\cite{mao2021accelerating}, our current results are simulation-only. 
Second, our method is computationally expensive: reconstructing a $128\times128$ image from 2,000 distorted observations takes around three hours on a single GPU (Nvidia RTX A6000). 

% \begin{figure*}[!t]
%     \centering
%     \includegraphics[width=1.0\textwidth]{figures/main_comparison.pdf}
%     \caption{{\bf Reconstructing Sharp Images from Blurry Measurements}. Given the reference images (a), we simulate 2000 observations distorted by spatially-varying turbulence. Typical distorted frames are shown in (b) and handpicked lucky frames are shown in (c). The results produced by~\cite{aittala2018burst} are shown in (d); they contain significant warping artifacts. The results produced by~\cite{Mao2020Turbulence} are shown in (e); they avoid warping artifacts but are not much sharper than the lucky frames. Our method (f) produces sharp reconstructions whose fine details and structure are consistent with the reference images.} \label{fig:main_comparison}
% \end{figure*}

\section{Related Work}
\label{RelatedWork}
%\textcolor{blue}{(2) Existing turbulence methods don't take advantage of GANs. GANS have been used for inverse problems with great success. Turbulence is harder than problems studied in previous papers and existing methods can't handle turbulence because we have unknown forward model characterized by many latents.}

\subsection{Imaging Through Turbulence}
Over the past three decades, many multiframe blind deconvolution and turbulence-removal algorithms have been developed.
The oldest and most widely used approach to multiframe imaging through turbulence is lucky imaging. 
Lucky imaging is founded on the concept that even under moderately severe turbulence conditions, given a large number of measurements, there is a high probability that at least one measurement will experience very little turbulence-induced distortion~\cite{Fried1978ProbabilityOG}. 
Likewise, modern anisoplanatic turbulence removal algorithms~\cite{Mao2020Turbulence}  often extract the lucky patches from an image sequence as an important step during the registration process. 
Nonetheless, lucky imaging provides limited benefits when all the measurements are blurry and there are no lucky frames to take advantage of.

When all the images in the sequence are blurry, one likely needs to perform deconvolution explicitly.
Expectation-maximization (EM) based deconvolution algorithms reconstruct the image while marginalizing over the distribution of the blur kernels~\cite{schulz1993multiframe}.
However, the computation cost of EM-based methods scales exponentially with the number of latent variables, making it impractical to deploy in applications involving high-dimensional continuous latent variables (i.e., high-resolution spatially-varying blur kernels).

A simpler alternative to EM is alternating minimization (AM).
AM-based deconvolution algorithms attempt to reconstruct both the blur kernels and the image by imposing various priors on each.
For instance, in~\cite{sroubek2011robust} the authors impose a sparse total variation prior on the image and a sparse and positive constraint on the blur kernels.
Recent efforts have imposed more accurate models on the images (non-local self-similarity~\cite{dabov2007image}) and the blur kernels (Kolmogorov turbulence~\cite{Mao2020Turbulence}).
To our knowledge, the method developed in~\cite{Mao2020Turbulence} represents the current state-of-the-art in multiframe imaging through turbulence.

Deep learning provides the possibility to develop turbulence removal algorithms that benefit from a large amount of training data.
Several recent works have trained deep neural networks to remove atmospheric distortions from a single image~\cite{yasarla2020learning, yasarla2021learning, Lau2020ATFaceGANSF}.
Recently, Mao et al.~\cite{mao2021accelerating} trained a convolutional neural network to reconstruct a sharp image from $50$ frames degraded by atmospheric turbulence. 
Their method achieved performance only incrementally worse than state-of-the-art classical methods while running significantly faster. 
For more traditional blind deconvolution tasks, like removing camera shake, researchers have developed methods~\cite{wieschollek2017learning, aittala2018burst} that generalize across an arbitrary number of frames. 
All these methods train their network to a specific blur kernel family or turbulence level, limiting their adaptability.

\vspace{-10pt}
\subsection{Imaging Inverse Problem with Generative Networks}
\label{ssec:InversewGAN}
By constraining reconstructed images to lie within the range of a deep generative network and then finding the latent parameters consistent with the measurements, several research groups have solved under-determined imaging inverse problems such as compressive sensing~\cite{bora2017compressed, wu2019deep, latorre2019fast,whang2020compressed,jalal2020robust,daras2021intermediate} and phase retrieval~\cite{hand2018phase, liu2021towards,metzler2021deep}. 
As shown in Deep Image Prior (DIP)~\cite{Ulyanov2018DeepIP}, a similar framework can be effective even if the generative network is untrained.

By estimating both the object and unknown forward model, this framework can be adapted to problems with unknown forward models, such as blind demodulation~\cite{hand2019global}, blind deconvolution~\cite{ren2020neural, asim2020blind}, phase retrieval with optical aberrations~\cite{Bostan:20}, and blind super-resolution~\cite{liang2021flow}.
However, such approaches, which largely boil down to alternating minimization with respect to both the scene's and forward model's latent spaces, do not easily apply to our multiframe anisoplanatic turbulence removal problem because they would require reconstructing all the blur kernels associated with all the pixels of all the observations.

\subsection{CryoGAN}
CryoGAN~\cite{Gupta2021CryoGAN} similarly reconstructs images by fitting a network to a collection of measurements, but it does so using a fundamentally different approach which compares the {\em  distributions} of {simulated} and {real} observations. We refer to this approach as adversarial sensing.%All the aforementioned methods also compare simulated and real observations, so I wouldn't emphasize those words

In CryoGAN, adversarial sensing compares the distributions of {simulated} and {real} tomographic projections and updates the underlying reconstruction so that those two distributions become more homogeneous.
Since the network is fitted over entire distributions, not individual observations or predictions, adversarial sensing allows one to perform reconstruction even when one-to-one correspondence between an observation and its associated latent variables cannot be established.
In the context of cryo-EM, this challenge arises because the latent orientation associated with each observed projection of the 3D object (molecule) is highly under-determined.
In this work, we extend adversarial sensing to the task of turbulence removal.
\section{Methods: Adversarial Sensing}
\label{sec:Methods}
\label{Methods}
%\textcolor{blue}{(3) Our method: Adversarial sensing (as proposed in cryoGAN) allows you to deal with such situations. Theorem goes here somewhere.} 
Our goal is to recover the underlying target scene $\pmb{x}$ given a set of observations $\{\pmb{y}_{i}\}_{i=1}^L$, characterized by the equation
\begin{equation}
%\pmb{y}_{i}(u,v) = \pmb{h}_{i}(u,v) * \pmb{x} + \pmb{\epsilon}_{i},
\pmb{y}_i[u,v] = (\pmb{h}_i(u,v) * \pmb{x})[u,v] + \pmb{\epsilon}_i[u,v],
\label{eq:2}
\end{equation}
where the set of spatially-varying blur kernels $\{\pmb{h}_{i}\}_{i=1}^L$ are unknown. We approach this challenging blind deconvolution problem with adversarial sensing.

%\textcolor{red}{The below description of our network needs to be modified to accurately describe both the DIP and stylegan based approach}

Our framework is illustrated in Fig.~\ref{fig:teaser} and described in Algorithm~\ref{alg:turbugan}. 
An either pre-trained~\cite{karras2020analyzing,Sauer2022StyleGANXLSS} or randomly initialized~\cite{Ulyanov2018DeepIP} generator network $\pmb{\mathcal{G}}_{\pmb{\theta}}$ takes in a fixed input $\pmb{z}$ and produce a single image $\tilde{\pmb{x}}$, which serves as the estimated reconstruction of the distortion-free scene. 
This reconstruction is fed to a physics-based differentiable forward model which synthesizes distorted measurements, which we denote with $\pmb{h}* \tilde{\pmb{x}}$. 
These synthesized measurements are then sent to a discriminator, which is trained to differentiate these synthetic measurements from the real observations, which we denote with $\pmb{h}* {\pmb{x}}$. 
In effect, the discriminator provides a loss on how close the synthesized data's distribution, $p_{h* \tilde{x}}(\pmb{y})$, is to the true data's distribution, $p_{h* x}(\pmb{y})$. 
The discriminator loss is used to update the generator network's weights $\pmb{\theta}$ so as to improve its reconstruction $\tilde{\pmb{x}}$ until $p_{h* \tilde{x}}(\pmb{y})$ and  $p_{h* x}(\pmb{y})$ are indistinguishable.
We may optionally back-propagate the loss to the differentiable forward model and update its parameters as well.

\begin{algorithm}[t!]
	\caption{\edit{TurbuGAN}} 
 % \text{\textcolor{blue}{~STILL NEED TO MENTION THAT WE REQUIRE FORWARD MODEL AS AN INPUT}}
	\label{alg:turbugan}
	\begin{algorithmic}[1]

		\renewcommand{\algorithmicrequire}{\textbf{Input:}}
		\REQUIRE $\text{(1) Turbulence-distorted observations} {\{\pmb{y}_i\}}_{i = 1}^{K},$\newline\text{\quad~\thinspace}(2) two regularization parameter $p_d$ and $p_r$,\newline \text{\quad~\thinspace}(3) an initial estimate of the turbulence aperture and Fried parameter ratio $D/r_0$, \newline \text{\quad~\thinspace}(4) a fast differentiable random turbulence simulator $\pmb{\mathcal{T}}(\cdot,D/r_0)$, such as~\cite{mao2021accelerating}.
		% \STATE \edit{$\pmb{x} \leftarrow \text{DirectInversion}_{\widehat{\pmb{\theta}}}(\pmb{y})$}
        \STATE \text{Initialize the scene estimate}  ${\tilde{\pmb{x}}}_{init}$ with either the mean of $\{\pmb{y}_i\}$ or the reconstruction from \cite{Mao2020Turbulence}
        %\STATE \text{~~~~~~~~~~~~~~~~~~~~~~~~~~~~~~~~~~~~~either the mean of $\{\pmb{y}_i\}$, or the reconstruction from \cite{Mao2020Turbulence}} 
        \IF{we use an \pmb{untrained} generator $\pmb{\mathcal{G}}$ like DIP~\cite{Ulyanov2018DeepIP}}
        \STATE\text{{Initialize $\pmb{\mathcal{G}}$'s weights such that $\pmb{\mathcal{G}}(\pmb{z})$ matches $\tilde{\pmb{x}}_{init}$, where $\pmb{z}$ is a fixed random input vector}}
        \ELSIF{we use a \pmb{pretrained} generator $\pmb{\mathcal{G}}$ like StyleGAN~\cite{Sauer2022StyleGANXLSS}}
        \STATE\text{{Fix $\pmb{\mathcal{G}}$'s weights and initialize its input $\pmb{z}$ such that $\pmb{\mathcal{G}}(\pmb{z})$ matches $\tilde{\pmb{x}}_{init}$ through PTI \cite{roich2021pivotal}}}
        \ENDIF
        % \STATE\text{{Initialize the generator $\pmb{\mathcal{G}}$ and its input $\pmb{z}$ such that $\pmb{\mathcal{G}}(\pmb{z})$ matches with $\tilde{\pmb{x}}_{init}$ through PTI \cite{roich2021pivotal}}}
        \STATE \text{Fix  $\pmb{\mathcal{G}}$'s input vector $\pmb{z}$ throughout training}

        %\STATE \text{Initialize the generator  and then fix it in later training process } $\pmb{z} = {\arg\min}_{\pmb{z}} ||\pmb{\mathcal{G}}(\pmb{z}) - {\tilde{\pmb{x}}_{init}}||^{2}_{2} $ 
        \STATE \text{}$ $
		\FOR {$k = 1,2,... $}          
            % \STATE \text{Use the turbulence simulator $\pmb{g}(\cdot)$ to randomly sample a set of spatially-varying blur kernels $\{\pmb{\tilde{h}}_{i}\}_{i = 1}^{K}$,} for which the details are described in Section~\ref{ssec:Zern}. $ $
            % \STATE \text{}$ $
            \STATE\text{\textbf{Update the discriminator:}}$ $
            \STATE \text{Freeze generator $\pmb{\mathcal{G}}$ and unfreeze discriminator $\pmb{D}$}
            \STATE $ \text{Get current scene estimate } \pmb{\tilde{x}} = \pmb{\mathcal{G}}(\pmb{z}) $
            \STATE \text{Randomly render fake observations} using the fast turbulence simulator $\{\pmb{\tilde{y}}_{i}\}_{i = 1}^{K} = \pmb{\mathcal{T}}(\pmb{\tilde{x}}, D/r_0)$
            % \STATE \text{Render fake observations } \{$\pmb{\tilde{y}}_{i}(u,v)\}_{i = 1}^{K} = \{\pmb{\tilde{h}}_{i}(u,v)\}_{i = 1}^{K} * \pmb{\tilde{x}}$.
            \STATE \text{Compute loss for discriminator $\pmb{D}$:}$ $
            \STATE \text{~~~~}$L^{real}_{D} = p_d \frac{1}{K} \sum_{i=1}^{K} \pmb{D}(\pmb{y}_i)^2 - \frac{1}{K} \sum_{i=1}^{K} \pmb{D}(\pmb{y}_i)$
            \STATE \text{~~~~}$L^{fake}_{D} = \frac{1}{K} \sum_{i=1}^{K} \pmb{D}(\pmb{\tilde{y}}_i)$
            \STATE \text{~~~~}$L^{mix}_{D} = p_r \frac{1}{K} \sum_{i=1}^{K}(\| \nabla_{\pmb{b}} \pmb{D}(\pmb{b}) \|_2 - 1)^{2}$ with $\pmb{b} = \alpha \pmb{ \tilde{y} }_i + (1 - \alpha) \pmb{y}_i$ and $\alpha \sim U[0, 1]$ %\text{\# {\it Lipschitz constraint}}
            \STATE \text{~~~} $L_{D} = L^{real}_{D} + L^{fake}_{D} + L^{mix}_{D}$
            \STATE \text{Backpropagate $L_D$ to update  discriminator $\pmb{D}$'s} weights
            \STATE \text{}$ $
            \STATE\text{\textbf{Update the generator:}}$ $
            \STATE \text{Unfreeze generator $\pmb{\mathcal{G}}$ and freeze discriminator $\pmb{D}$}
            \STATE $ \text{Get current scene estimate } \pmb{\tilde{x}} = \pmb{\mathcal{G}}(\pmb{z}) $
            \STATE \text{Randomly render fake observations} using the fast turbulence simulator $\{\pmb{\tilde{y}}_{i}\}_{i = 1}^{K} = \pmb{\mathcal{T}}(\pmb{\tilde{x}}, D/r_0)$
            % \STATE \text{Render fake observations } \{$\pmb{\tilde{y}}_{i}\}_{i = 1}^{K}$ from $\pmb{\tilde{x}}$ using the fast turbulence simulator $\pmb{\mathcal{T}}(\cdot)$.
            % \STATE \text{Render fake observations } $\{\pmb{\tilde{y}}_{i}(u,v)\}_{i = 1}^{K} = \{\pmb{\tilde{h}}_{i}(u,v)\}_{i = 1}^{K} * \pmb{\tilde{x}}$.
            \STATE \text{Compute loss for generator $\pmb{\mathcal{G}}$:}$ $
            \STATE \text{~~~~}$L_{\mathcal{G}} = \frac{1}{K} \sum_{i=1}^{K} \pmb{D}(\pmb{\tilde{y}}_i)$
            \STATE \text{Backpropagate $L_\mathcal{G}$ to update generator $\pmb{\mathcal{G}}$'s} weights and the estimated turbulence parameter $D/r_0$
            \STATE \text{}$ $
		\ENDFOR
      \renewcommand{\algorithmicensure}{\textbf{Output:}} \ENSURE $\pmb{\tilde{x}}$
	\end{algorithmic}
\end{algorithm}

\subsection{Physically Accurate Differentiable Forward Model for Anisoplanatic Atmospheric Turbulence}\label{ssec:Zern}

%\subsection{Physically Accurate Differentiable Forward Model for Anisoplanatic Atmospheric Turbulence}\label{ssec:Zern}
%Our adversarial sensing framework requires a representation of the scene to facilitate training and reconstruction. Instead of directly optimizing $\pmb{x} \in \mathbb{R}^{{N}\times{N}}$ as a 2D matrix, we optimize the weights $\pmb{\theta}$ of a convolutional neural network $\pmb{\mathcal{G}}_{\pmb{\theta}}$ where $x=\pmb{\mathcal{G}}_{\pmb{\theta}}(z)$.  As such, the neural network $\pmb{\mathcal{G}}_{\pmb{\theta}}$ essentially serves as the representation of the scene $\pmb{x}$. As explained in Section~\ref{ssec:AdversarialTraining}, during training, $z$ is fixed after initialization while a discriminator provides the gradients to update $\pmb{\theta}$. The specific initialization scheme of $z$ varies based on the level of knowledge we have about the target scene, with more details in Section~\ref{Experiments}. 

% Moreover, 
Our framework relies upon having access to a physically accurate forward model that can simulate realistic turbulence on top of the estimate of $\pmb{x}$, so that the synthetic measurements follow the same distribution as the real measurements.
This forward model also needs to be differentiable to enable loss back-propagation to update the generator network $\pmb{\mathcal{G}}$.

As explained in~\cite{GoodmanFourierOptics}, the blur kernel $\pmb{h}$ associated with the turbulence at location $(u,v)$ is related to the phase errors $\pmb{\phi}(u,v)$ at that location via the expression
\begin{align}\label{eq:PhasetoPSF}
\pmb{h}(u,v) =|\mathcal{F}(e^{-j\pmb{\phi}(u,v)})|^2,
\end{align}
\edit{where $\mathcal{F}$ denotes the 2D Fourier transform. Like the spatially varying blur kernel $\pmb{h}(u,v)$, the phase error $\pmb{\phi} \colon (u,v)\in\mathbb{R}^{N\times N} \mapsto \pmb{\phi}(u,v)\in\mathbb{R}^{M\times M}$ is a bivariate matrix-valued function.}

This relationship allows one to generate realistic blur kernels by generating realistic phase errors $\pmb{\phi}(u,v)$. % for all $(u,v)$.
Specifically, at each pixel $(u,v)$ one can approximate $\pmb{\phi}(u,v)$ with the first $M$ Zernike polynomials as
\begin{equation}
\pmb{\phi}(u,v) = \sum_{i=1}^{M} \alpha_{i} Z_{i}(u,v),
\label{eq:weighted_zernike}
\end{equation}
where $Z_i$ is the $i^{th}$ Zernike basis and $\alpha_i$ is its corresponding coefficient~\cite{Noll1976ZernikePA}.
The vector of coefficients $\pmb{\alpha}$ follows a zero-mean normal distribution with covariance matrix $\mathbf{\Sigma}$. 
% \begin{align}\label{eq:phasecovariance}
%   \alpha_{i} \sim \mathcal{N}(0,\,\mathbf{\Sigma}).
% \end{align}
Detailed derivations~\cite{Noll1976ZernikePA, chimitt2020simulating} show that $\mathbf{\Sigma}$ depends on the ratio of the imaging system's aperture's diameter $D$ and the Fried parameter $r_{0}$~\cite{Fried1978ProbabilityOG}, which characterizes the turbulence strength.
% Therefore, we may generate physically realistic blur kernels through differentiable sampling from Equations \eqref{eq:PhasetoPSF}, \eqref{eq:weighted_zernike}, \eqref{eq:phasecovariance}.

%%%%%%%%%%%%%%%%%%%%%%%
The correlations between the Zernike phase errors at adjacent locations are similarly well understood~\cite{chimitt2020simulating}. 
Thus, to model propagation through atmospheric turbulence, one can (A) sample a collection of correlated Zernike coefficients for every image pixel, (B) use Equations \eqref{eq:PhasetoPSF} and \eqref{eq:weighted_zernike} to compute the PSF associated with each pixel, and (C) convolve each pixel in the scene with its associated PSF~\cite{chimitt2020simulating}. 
As one might expect, performing these computations explicitly is extremely computationally expensive.

Fortunately, the recently proposed phase-to-space (P2S) transform~\cite{mao2021accelerating} bypasses the computationally expensive blur kernel generation process (Equations \eqref{eq:PhasetoPSF} and \eqref{eq:weighted_zernike}) by training a neural network to learn a mapping directly between the Zernike coefficients and blur kernels. 
Simulation results demonstrate that the blur kernels generated with the P2S transform are physically accurate, consistent with existing theory, and can be evaluated $300$--$1000\times$ faster than previous physics-based forward models~\cite{mao2021accelerating}.
Accordingly, we adopt the P2S transform both to generate the ground truth observations and serve as our differentiable rendering engine to generate the synthetic measurements during training.

\subsection{Optimization Objectives}\label{ssec:AdversarialTraining}
Adversarial sensing follows the standard adversarial learning paradigm, where one trains a generator $\pmb{\mathcal{G}}$ to trick a discriminator $\pmb{D}$  and also trains $\pmb{D}$ not to get tricked. 
However, different than conventional GAN training, the discriminator loss is computed based on the distorted measurements, rather than the direct output of the generator. \edit{During the optimization process, we update the weights of the generator and the discriminator and the tunable variables of the differentiable renderer. The generator's input latent code $\pmb{z}$ is fixed throughout training.}%, which is distortion-free.

\subsubsection{Optimizing the Discriminator}
To train the discriminator, for each training batch we first gather a collection of $K$ observed distorted measurements $\pmb{y}$.
Next, we obtain the current estimated target image $\pmb{\tilde{x}} = \pmb{\mathcal{G}}(z)$ and pass it through the differentiable foward model to construct fake observations
\{$\pmb{\tilde{y}}_{i}$\}.
%\{$\pmb{\tilde{y}}_{i}(u,v) = \pmb{\tilde{h}}_{i}(u,v) * \pmb{\tilde{x}}\}$.

We then send both \{$\pmb{\tilde{y}}_{i}$\} and \{$\pmb{{y}}_{i}$\} to the discriminator to compute \{$\pmb{D}(\pmb{y}_i)$\} and \{$\pmb{D}(\pmb{\tilde{y}}_i)$\}.
Finally, we update the discriminator $D$ with loss $L_{D} = L^{real}_{D} + L^{fake}_{D} + L^{mix}_{D}$, where the three terms are defined as:
\begin{equation}
    L^{real}_{D} = \frac{p_d}{K} \sum_{i=1}^{K} \pmb{D}(\pmb{y}_i)^2 - \frac{1}{K} \sum_{i=1}^{K} \pmb{D}(\pmb{y}_i),
\end{equation}

\begin{equation}
    L^{fake}_{D} = \frac{1}{K} \sum_{i=1}^{K} \pmb{D}(\pmb{\tilde{y}}_i).
\end{equation}

\begin{equation}
    L^{mix}_{D} = \frac{p_r}{K} \sum_{i=1}^{K}(\| \nabla_{\pmb{b}} \pmb{D}(\pmb{b}) \|_2 - 1)^{2}
\end{equation}

This loss function is based on the Wasserstein GAN loss~\cite{Gulrajani2017ImprovedTO}, and $p_d$ and $p_r$ are both regulatory parameters.
In the third term $L^{mix}_{D}$, $\pmb{b} = \alpha \pmb{ \tilde{y} }_i + (1 - \alpha) \pmb{y}_i$, where $\alpha$ are randomly sampled from $[0, 1]$.
This loss term softly enforces the Lipschitz constraint on the discriminator and decreases training difficulty.

\subsubsection{Optimizing the Generator} \label{sssec:proxy_generator}
Optimizing the generator is straightforward.
We encourage the generator to produce fake samples that are scored higher by the discriminator. The generator's loss is defined as:
\begin{equation}
    L_{\mathcal{G}} = -\frac{1}{K} \sum_{i=1}^{K} \pmb{D}(\pmb{\tilde{y}}_i).
\label{eq:7}
\end{equation}

\subsection{Theoretical Justification}
\label{ssec:theorem}
Before delving into the experimental results, we demonstrate that in the simpler case of isoplanatic turbulence (spatially-invariant blur) with a known distribution, matching the distributions of the simulated and observed noise-free measurements reconstructs the scene. %This provides hope that adversarial sensing can potentially reconstruct our captured scenes.
%we prove that in the case of isoplanar turbulence (spatially-invariant forward model), matching the distributions of the simulated and observed measurements reconstructs the scene.
% We now provide additional details about each step of this process.
\begin{theorem}
\label{thm:1}
Let $p_{\pmb{h}* \pmb{x}}(\pmb{y})$ denote the distribution associated with noiseless spatially-invariant measurements
\begin{align}
\pmb{y}=\pmb{h}*\pmb{x},~\text{where}~\pmb{h}\sim p_{\pmb{h}}.
\end{align}
Let $p_{h* \tilde{x}}(\pmb{y})$ denote the distribution associated with measurements formed by a reconstruction $\tilde{\pmb{x}}$ and blur kernels following $p_h$. 
Then,%, assuming $\pmb{x}_o$ and $\tilde{\pmb{x}}$ are  non-negative, 
\begin{align}\label{eqn:MainTheorem}
p_{\pmb{h}* \pmb{x}}(\pmb{y})=p_{\pmb{h}* \tilde{\pmb{x}}}(\pmb{y})\implies \pmb{S}\circ|\mathcal{F}\pmb{x}|=\pmb{S}\circ|\mathcal{F}\tilde{\pmb{x}}|,
\end{align}
where $\circ$ denotes a Hadamard (elementwise) product, $\mathcal{F}$ denotes the 2D Fourier transform, and $\pmb{S}=\mathbbm{1}_{+}(\mathbb{E}[|\mathcal{F}\pmb{h}|^2])$ where $\mathbbm{1}_{+}(\cdot)$ denotes an indicator function that is $1$ if its argument is greater than 0.% That is $g$ is a filter whose frequency response is $0$ at frequencies where the power spectral density of $h$ is $0$ and $1$ and at all other frequencies. 
% $(\mathcal{F}h)(k)=0$ for all $h\sim p_h$ and 1 for all other frequencies.
\end{theorem}
In effect, Theorem~\ref{thm:1} states that if adversarial sensing can match the distributions of the observed and simulated measurements, it will reconstruct $\pmb{x}$ accurately up to the invariances (\eg~translation)~associated with the masked phase retrieval problem defined by the right-hand side of Equation~\eqref{eqn:MainTheorem}~\edit{\cite{bendory2017fourier}}.
%The aforementioned mask is defined by the frequencies where the power spectral density of the stochastic blur kernel is non-zero. 
%A proof can be found in the Appendix.%~\ref{sec:appendix}. 
% ($\mathbb{E}[|\mathcal{F}\pmb{h}|^2]$)

\begin{custom_corollary}
\label{corr:1}
If one further makes the (strong) assumption that $\mathbb{E}[\mathcal{F}\pmb{h}](K_u,K_v)\neq 0$ for all spatial frequencies $(K_u,K_v)$, then $\pmb{x}=\tilde{\pmb{x}}$.  
\end{custom_corollary}
\begin{custom_corollary}

\label{corr:2}
If one reconstructs $\pmb{x}$ using a misspecified forward model $\pmb{h}_s\neq \pmb{h}$, then assuming that $\pmb{h}_s=\pmb{h}_{correction}*\pmb{h}$ for all $\pmb{h}\sim p_{\pmb{h}}$ and that $\mathbb{E}[\mathcal{F}\pmb{h}_{correction}](K_u,K_v)\neq 0$ and $\mathbb{E}[\mathcal{F}\pmb{h}](K_u,K_v)\neq 0$ for all spatial frequencies $(K_u,K_v)$, one will reconstruct a filtered version of the signal, $\tilde{\pmb{x}}=\pmb{h}_{correction}*\pmb{x}$.
\end{custom_corollary}

Proofs of Theorem~\ref{thm:1}, Corollary~\ref{corr:1}, and Corollary~\ref{corr:2} can be found in the Appendix.
\section{Experiments}\label{Experiments}
\begin{figure*}[t]
    \centering
    \SevenColSubfig{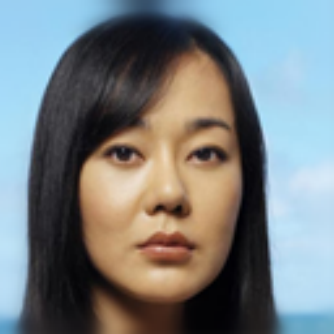}%
    \SevenColSubfig{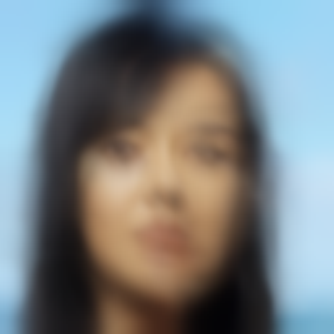}%
    \SevenColSubfig{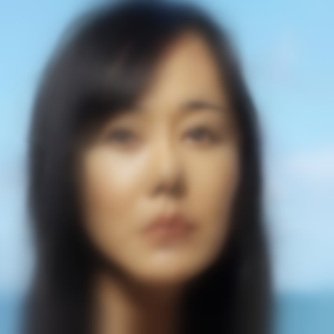}%
    \SevenColSubfig{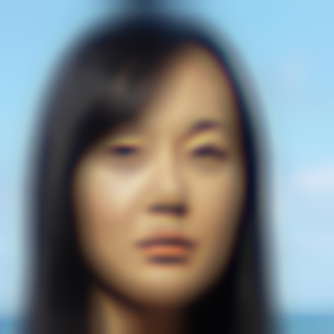}%
    \SevenColSubfig{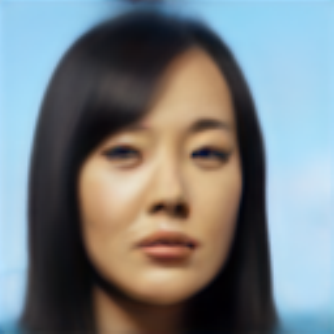}%
    \SevenColSubfig{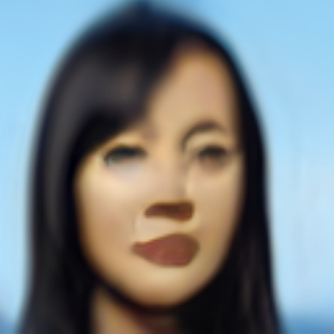}%
    \SevenColSubfig{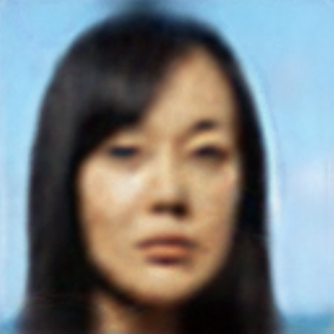}%

    \SevenColSubfig{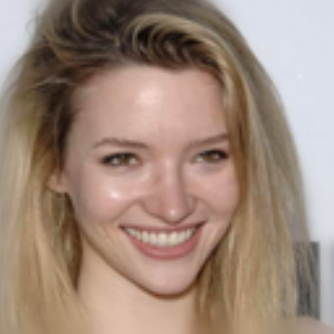}%
    \SevenColSubfig{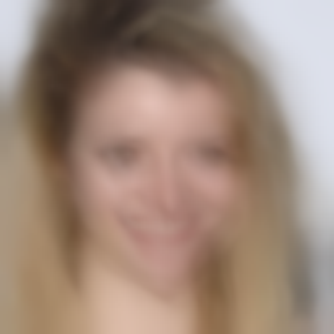}%
    \SevenColSubfig{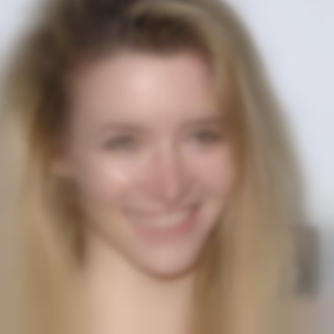}%
    \SevenColSubfig{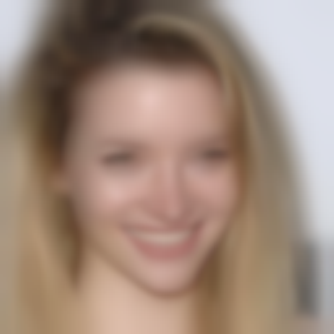}%
    \SevenColSubfig{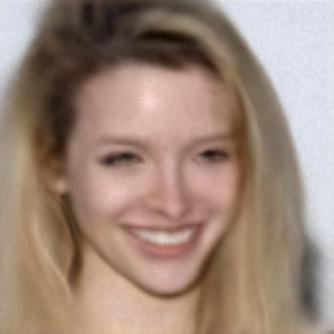}%
    \SevenColSubfig{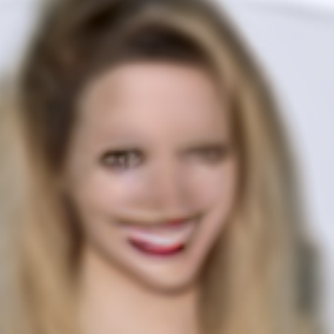}%
    \SevenColSubfig{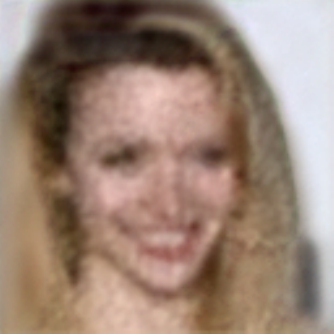}%

    \SevenColSubfig{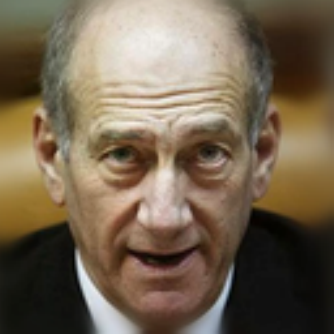}%
    \SevenColSubfig{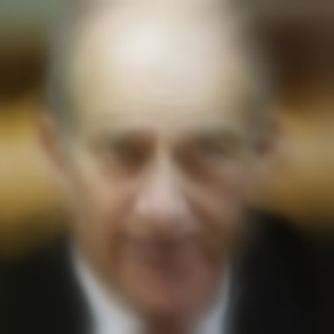}%
    \SevenColSubfig{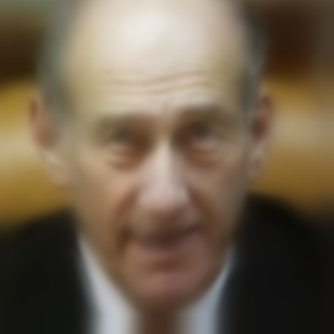}%
    \SevenColSubfig{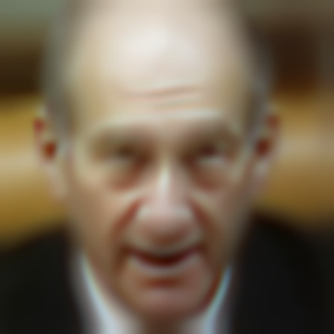}%
    \SevenColSubfig{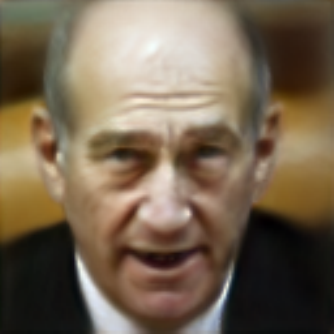}%
    \SevenColSubfig{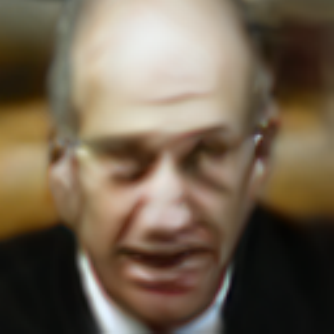}%
    \SevenColSubfig{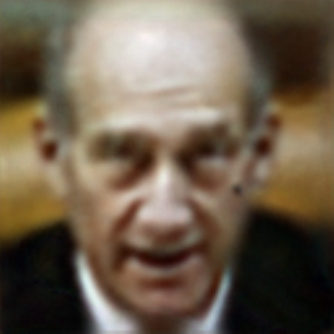}%

    \SevenColSubfigwCaption{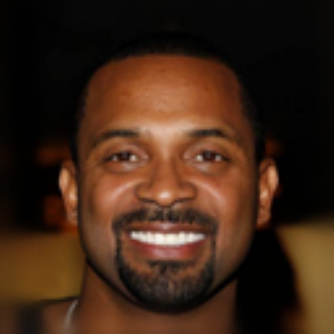}{Reference}%
    \SevenColSubfigwCaption{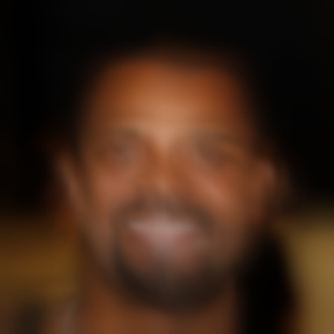}{Typical}%
    \SevenColSubfigwCaption{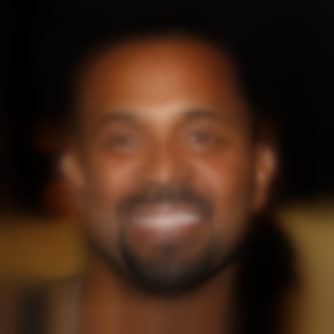}{Lucky}%
    \SevenColSubfigwCaption{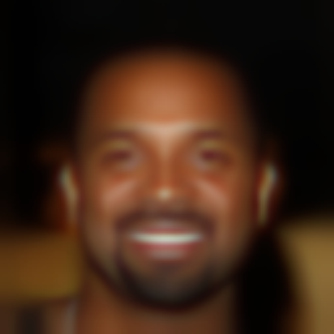}{Mao et al.}%
    \SevenColSubfigwCaption{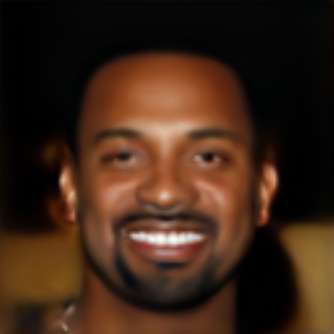}{Ours-Face}%
    \SevenColSubfigwCaption{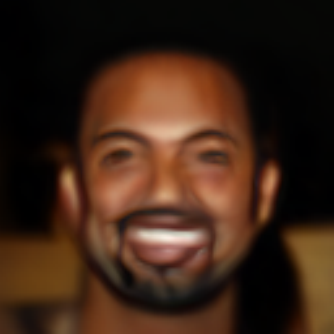}{Ours-IN}%
    \SevenColSubfigwCaption{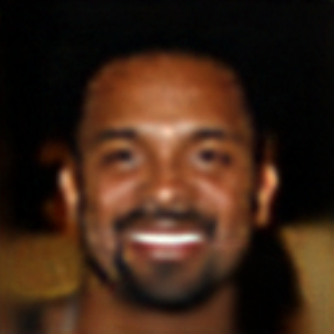}{Ours-DIP}%

    \caption{{\bf Reconstructions from Simulated Turbulence-Distorted Face Images.} Given the reference image (a), we simulate 2000 observations distorted by spatially-varying turbulence. Typical distorted frames are shown in (b) and handpicked lucky frames are shown in (c). The results produced by~\cite{Mao2020Turbulence} are shown in (d), which are not much sharper than the lucky frames. In (e)-(f), we show reconstructions produced by TurbuGAN under three conditions (details in~\ref{ssec:main_results}) with different levels of prior knowledge about the data domain (face images in this case). (e) utilize a StyleGAN generator pretrained on faces and show fine details and clear structure consistent with the reference images. (f) utilize a generator pretrained on natural images (ImageNet), while (g) assume no knowledge about the image domain by using a randomly initialized generator. Nonetheless, all conditions still achieve reasonable results, showing the flexibility of our framework in adapting to different levels of prior knowledge.} \label{fig:main_comparison}	%\vspace{-5pt}
\end{figure*}

\begin{figure*}[!ht]
    \centering

    \SevenColSubfig{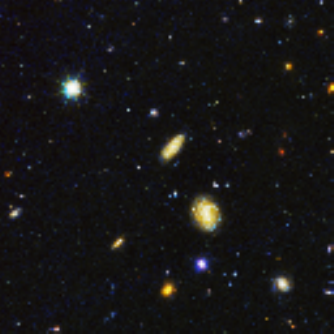}%
    \SevenColSubfig{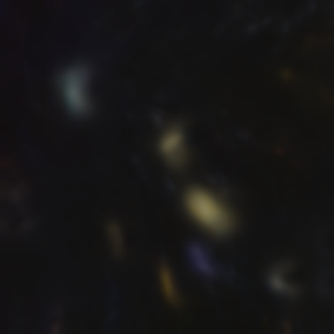}%
    \SevenColSubfig{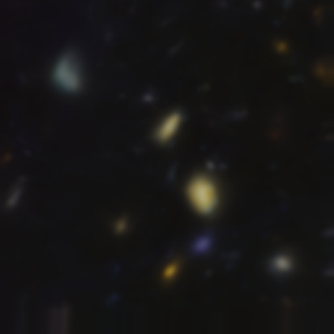}%
    \SevenColSubfig{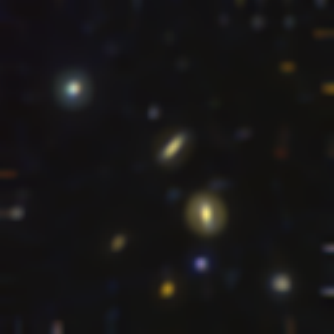}%
    \SevenColSubfig{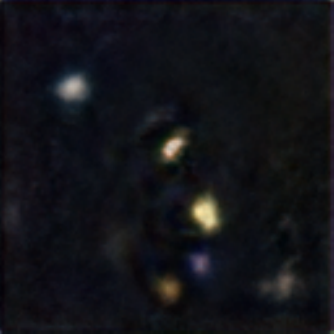}%
    \SevenColSubfig{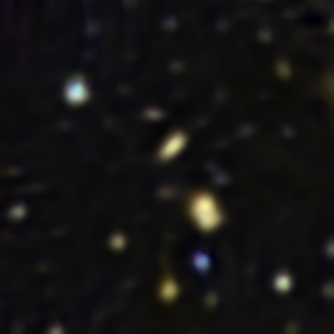}%
    \SevenColSubfig{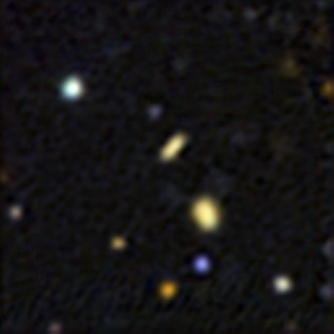}%

    \SevenColSubfig{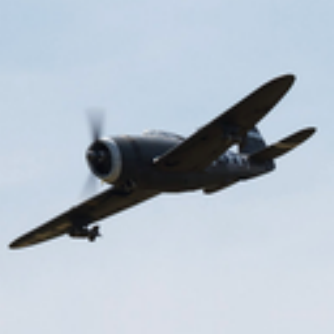}%
    \SevenColSubfig{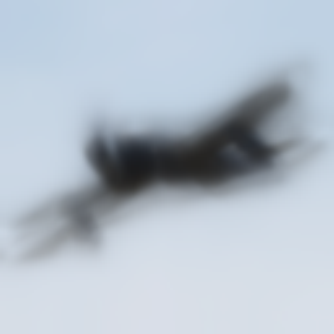}%
    \SevenColSubfig{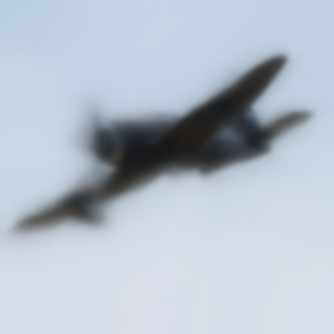}%
    \SevenColSubfig{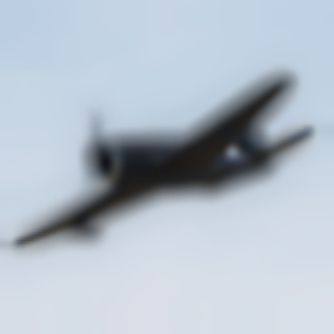}%
    \SevenColSubfig{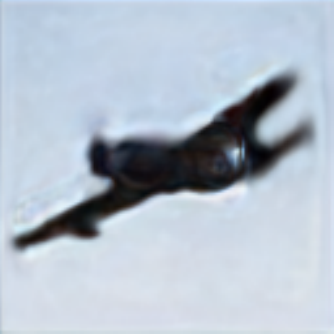}%
    \SevenColSubfig{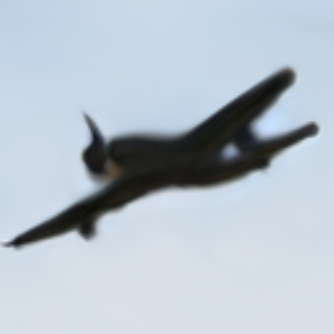}%
    \SevenColSubfig{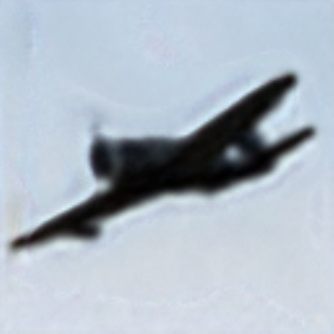}%
    
    \SevenColSubfig{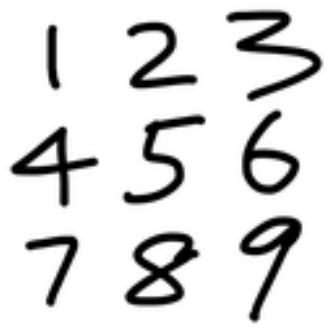}%
    \SevenColSubfig{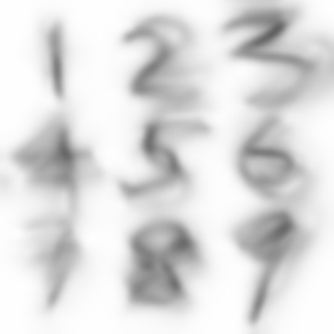}%
    \SevenColSubfig{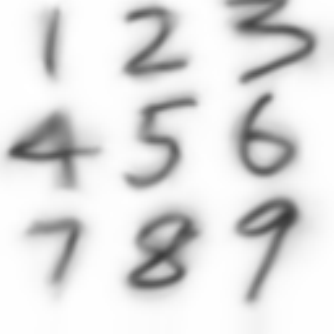}%
    \SevenColSubfig{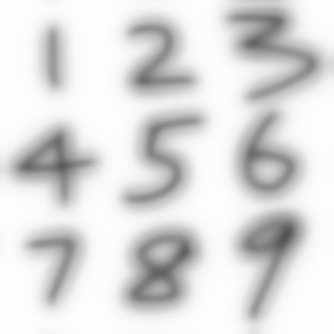}%
    \SevenColSubfig{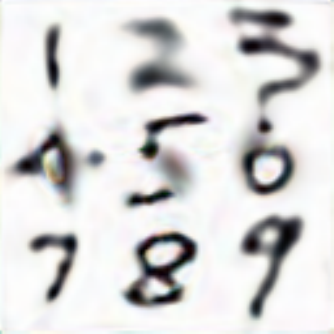}%
    \SevenColSubfig{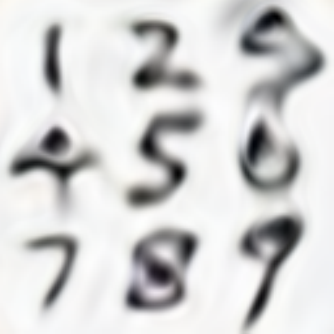}%
    \SevenColSubfig{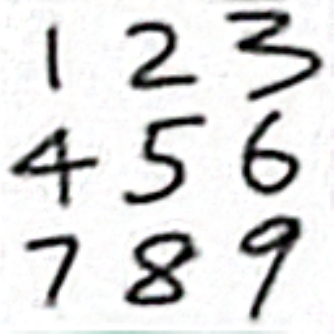}%

   \SevenColSubfig{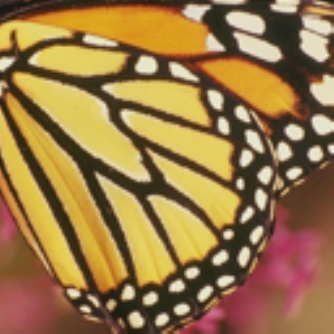}%
    \SevenColSubfig{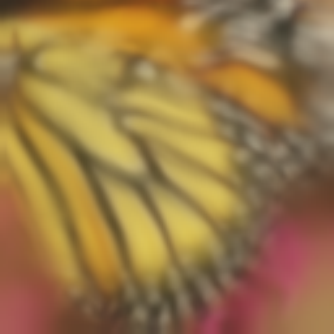}%
    \SevenColSubfig{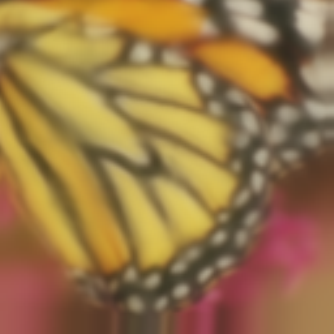}%
    \SevenColSubfig{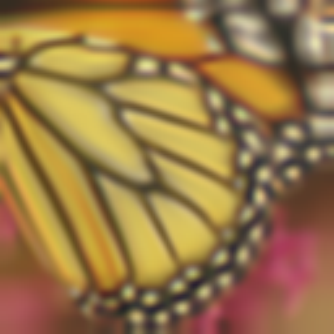}%
    \SevenColSubfig{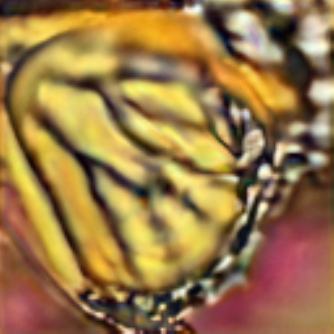}%
    % \SevenColSubfig{./figures/placeholder.png}%
    \SevenColSubfig{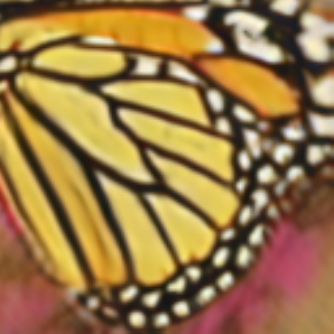}%
    \SevenColSubfig{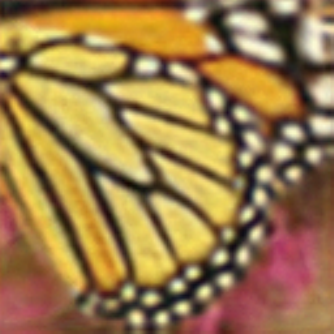}%
    % \SevenColSubfig{./figures/placeholder.png}%

    \SevenColSubfigwCaption{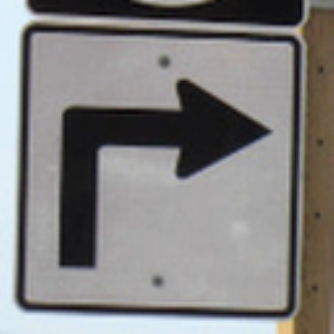}{Reference}%
    \SevenColSubfigwCaption{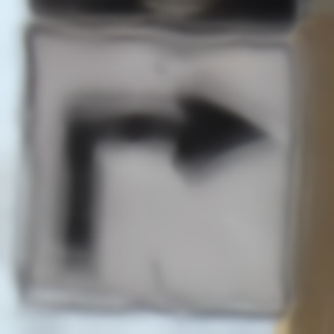}{Typical}%
    \SevenColSubfigwCaption{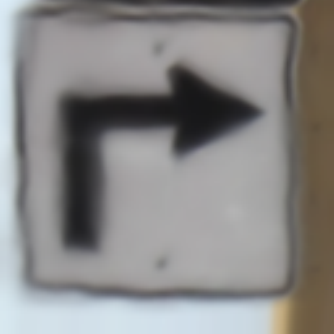}{Lucky}%
    \SevenColSubfigwCaption{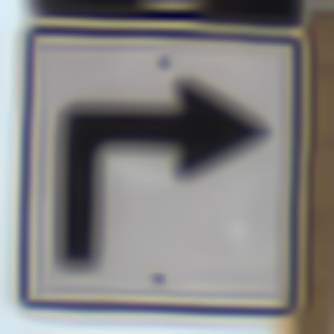}{Mao et al.}%
    \SevenColSubfigwCaption{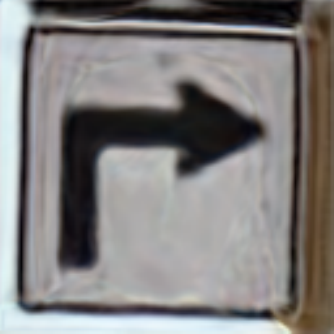}{Ours-Face}%
    \SevenColSubfigwCaption{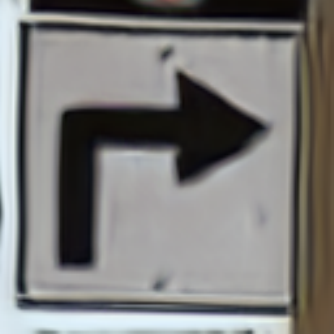}{Ours-IN}%
    \SevenColSubfigwCaption{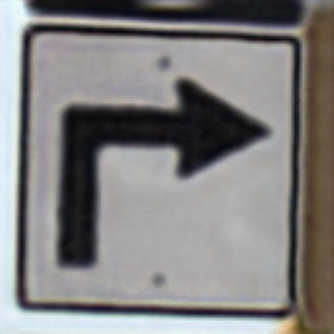}{Ours-DIP}%
    \caption{{\bf Reconstructions from Simulated Turbulence-Distorted General Images}. We extend to general, non-face images, with experiments done under turbulence strength $D/r_{0}=2.0$. The results produced by~\cite{Mao2020Turbulence}, shown in (d), contain relatively few artifacts but are generally blurry. When we initialize the generator with StyleGAN generator pretrained on face images, the results (e) show some visible artifacts since the initialized weights possess a strong prior on faces. We achieve better results (g) with generator pretrained on ImageNet, which has less specialized but more flexible prior over natural images. However, the digits results (third row) contain visible artifacts since they are not natural images and fall outside of the domain of ImageNet. When using a randomly initialized generator (DIP), the reconstructions (g) are sharp and contain minimal artifacts. These results show that TurbuGAN extends beyond face images and remains robust by flexibly adapting to appropriate levels of domain-specific prior knowledge.} \label{fig:imagenet}  
	%\vspace{-5pt}
\end{figure*}

\begin{figure}[!t]
    \centering
    \includegraphics[width=0.50\linewidth]{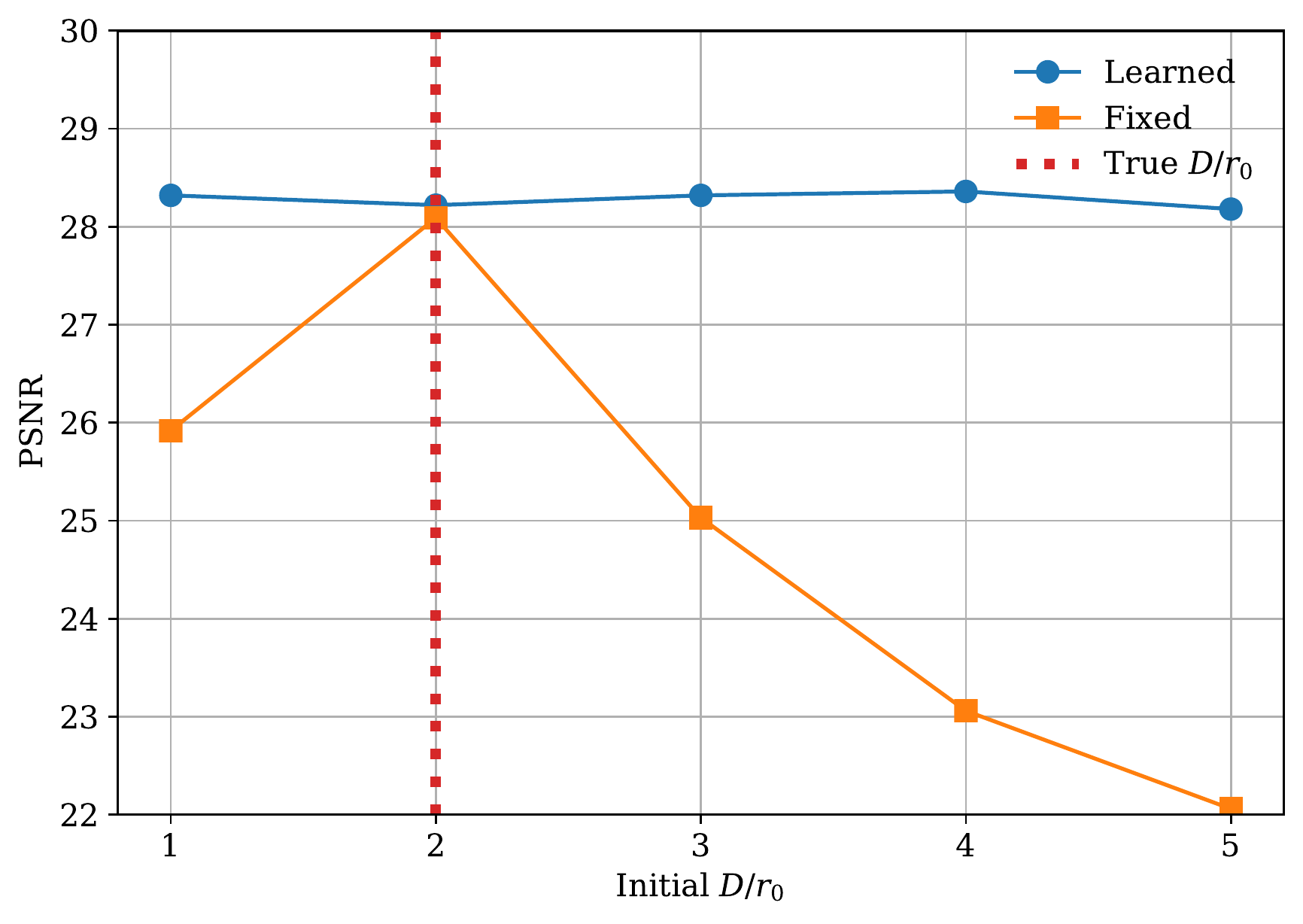}
    \caption{{\bf Adapting to Mismatched Turbulence Strength $D/r_{0}$.} We set the true unknown turbulence strength level $D/r_{0}$ equal to 2 (red vertical line). We initialize our framework with different $D/r_{0}$ levels ranging from 1 to 5, and we compare the performance (PSNR) between learning and fixing the initial $D/r_{0}$. The orange curve shows that the reconstruction quality significantly deteriorates when the framework sticks to a misspecified turbulence strength level. The blue curve shows that by updating $D/r_{0}$ during training, our method is able to overcome severe initial mismatches and achieve almost the same performance as when it is correctly initialized with $D/r_{0} = 2$. } \label{fig:learning_variances}
\end{figure}

In this section, we perform several experiments to evaluate the proposed method on the turbulence removal problem. 
During all TurbuGAN training, we apply the fast simulator introduced by Mao et al.~\cite{mao2021accelerating} to generate realistic anisoplanatic turbulence.
\edit{Unless otherwise noted, in all simulated experiments we use $L=2000$ measurements for each scene with an aperture diameter $D$ of 0.1 meters, a Fried parameter $r_{0}$ of 0.05 meters, and a distance-to-target of 1000 meters. The turbulence simulator assumes the central wavelength is 525 nm and the diffraction limit (which is a function of focal length and pixel pitch) is one pixel.} In each iteration, we use a batch size $K=32$ for both observed  and simulated measurements. 
All target images have a resolution of $128 \times 128$. Both $\pmb{D}$ and $\pmb{\mathcal{G}}$ are trained using the Adam optimizer~\cite{Kingma2015AdamAM}; we first warm up $\pmb{D}$ for 5000 iterations while fixing $\pmb{\mathcal{G}}$; afterwards, we alternate between updating $\pmb{D}$ for six iterations and $\pmb{\mathcal{G}}$ for one iteration. 
Training stops after 100,000 iterations and takes about three hours on an NVIDIA RTX A6000 GPU. The values of the two regulation parameters $p_d$ and $p_r$ are set to 0.001 and 10, respectively. Our implementation in PyTorch~\cite{torch2019} will be released upon acceptance.

{\bf Different Priors.} TurbuGAN allows us to flexibly select different strategies to initialize the neural network $\pmb{\mathcal{G}}_{\pmb{\theta}}$, each imposing different levels of prior knowledge about the scene.
In our experiments, we start with testing TurbuGAN on face images and then extend it to general, non-face images.
Therefore, we deploy TurbuGAN under three conditions: 1) {\it Face}, where we have the strongest level of prior knowledge about the domain of human faces, 2) {\it IN} (short for ImageNet), a moderate level of prior knowledge on the domain of natural images, and 3) {\it DIP}, no domain-specific knowledge.
For the {\it Face} condition, we use the weights of StyleGAN2~\cite{karras2020analyzing} pretrained on the FFHQ dataset~\cite{Karras2019ASG} containing real human face images.
For the {\it IN} condition, we use the weights of StyleGAN-XL~\cite{Sauer2022StyleGANXLSS} pretrained on the ImageNet dataset~\cite{Russakovsky2015ImageNetLS}.
For the {\it DIP} condition, we use PyTorch's default Kaiming initialization~\cite{He2015DelvingDI} to initialize the weights.

{\bf Initialization Schemes.} Before training under the first two conditions, we first optimize the input vector \textbf{$z$} similar to the first stage of the Pivotal Tuning Inversion (PTI) approach~\cite{roich2021pivotal}, which has demonstrated state-of-the-art performance on GAN inversion. 
Specifically, we optimize \textbf{$z$} such that the initial generator output matches the reconstructed result of~\cite{Mao2020Turbulence}. 
% When training under the third condition {\it DIP}, we first initialize the network output to be the mean of the distorted measurements.
We then optimize the network weights based on the training objective described in Section~\ref{ssec:AdversarialTraining}. 
\subsection{Results on Computationally Simulated Images}
\label{ssec:main_results}
We use the P2S turbulence simulator~\cite{mao2021accelerating} to computationally induce physically-accurate turbulence on images.
The selected images include real human faces from the CelebA dataset~\cite{liu2015faceattributes}, and non-face, publicly available images~\cite{Agustsson2017NTIRE2C, hille_2015} outside the ImageNet dataset~\cite{Russakovsky2015ImageNetLS}.
We compare TurbuGAN with a state-of-the-art turbulence-removal algorithm~\cite{Mao2020Turbulence} and show qualitative comparisons in Fig.~\ref{fig:main_comparison}.
TurbuGAN succeeds in restoring the original image, while the baseline method produces blurry images with obvious artifacts. 
Results in Fig.~\ref{fig:imagenet} further demonstrate that TurbuGAN extends to data domains where we cannot initialize the generator with domain-specific prior knowledge.

\subsection{Adapting to Misspecified Forward Models} \label{ssec:misspecified}
In the turbulence simulator~\cite{mao2021accelerating}, the turbulence strength is described by $D/r_{0}$, where $D$ is the aperture diameter and $r_{0}$ is the Fried parameter~\cite{Fried1978ProbabilityOG}. 
In previous experiments, we set the same $D/r_{0}$ values when we generate observed and simulated measurements.
However, while in real-world imaging scenarios the aperture diameter $D$ is usually known, the Fried parameter $r_{0}$ is likely unknown. 
Our framework would have limited use if it required knowing the precise turbulence strength $D/r_{0}$ of the scene.

\begin{figure*}[t]
    \centering
    % \SevenColSubfig{./figures/plane/plane-895b_128.pdf}%
    \FiveColSubfig{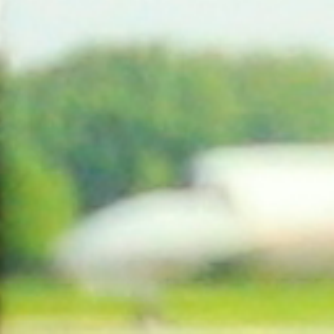}%
    \FiveColSubfig{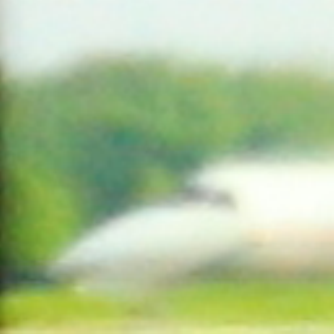}%
    \FiveColSubfig{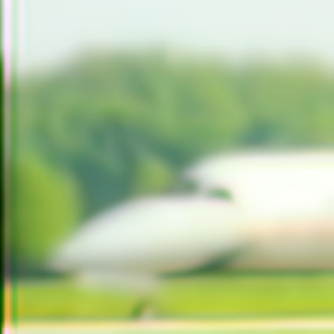}%
    % \SevenColSubfig{./figures/plane/plane-ours_24.91_Face.pdf}%
    \FiveColSubfig{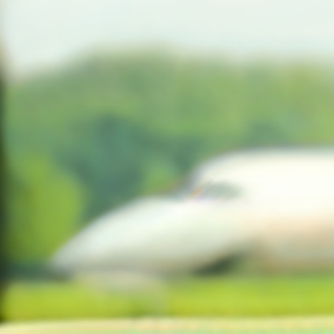}%
    \FiveColSubfig{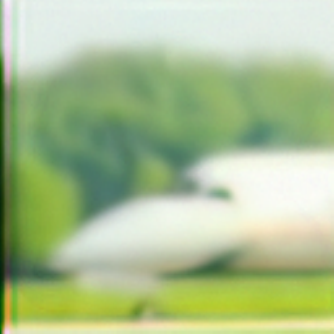}%

    \FiveColSubfigwCaption{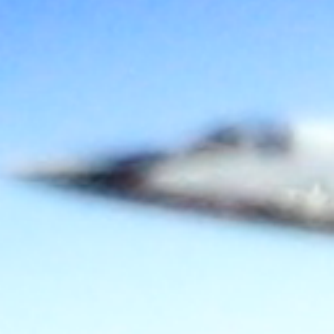}{Typical}%
    \FiveColSubfigwCaption{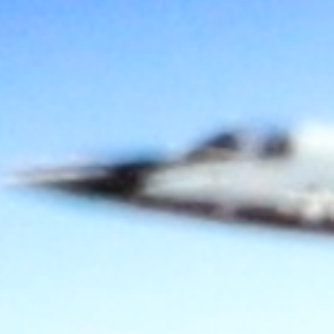}{Lucky}%
    \FiveColSubfigwCaption{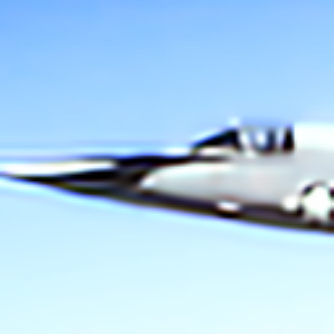}{Mao et al.}%
    \FiveColSubfigwCaption{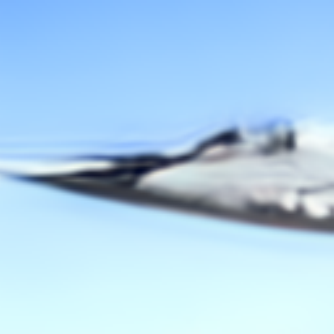}{Ours-IN}%
    \FiveColSubfigwCaption{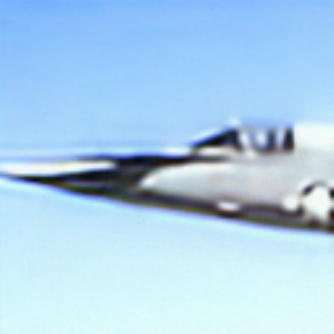}{Ours-DIP}%
    \caption{{\bf Reconstructions from Experimentally-Captured Turbulence-Distorted Images}. We further apply TurbuGAN to measurements distorted by real-world atmospheric turbulence. Each scene contains 100 measurements with unknown levels of turbulence distortion. Results show that TurbuGAN's performance is robust on real-world data. Our reconstructions when not assuming any domain-specific priors ({\it DIP}) appear sharper than when leveraging the prior from the ImageNet ({\it IN}) domain, as shown in (d) and (e).} \label{fig:realdata}
	%\vspace{-5pt}
\end{figure*}

\begin{figure}[th!]
    \vspace{1.2em}
    \centering
    %\captionsetup[subfloat]{farskip=2pt,captionskip=1pt}
    \subfloat[Typical Measurement]{\begin{overpic}[width=0.5\linewidth]{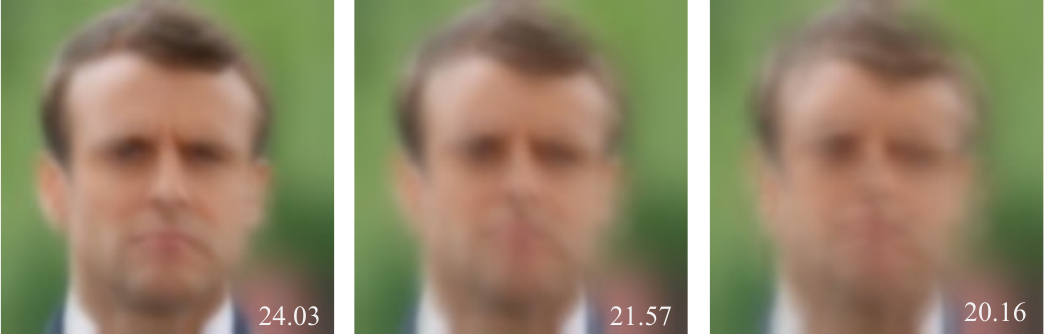}
      \put(7,33.5){$D/r_{0}=1$} 
      \put(39.5,33.5){$D/r_{0}=2$}
      \put(74,33.5){$D/r_{0}=3$}
    %   \put(10,33.5){$D/r_{0}=1$} 
    %   \put(43.5,33.5){$D/r_{0}=2$}
    %   \put(78,33.5){$D/r_{0}=3$}
      \end{overpic} \vspace{-4pt}}\\
    \subfloat[Mao et al.]{\includegraphics[width = 0.5\linewidth]{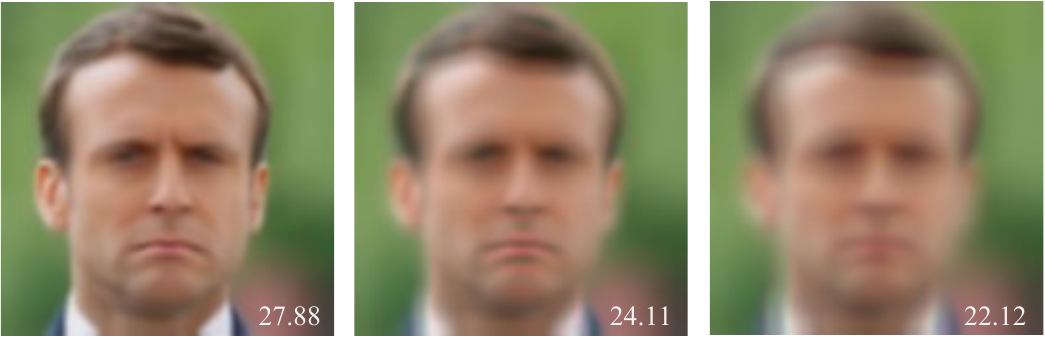} \vspace{-4pt}}\\
    \subfloat[Ours]{\includegraphics[width = 0.5\linewidth]{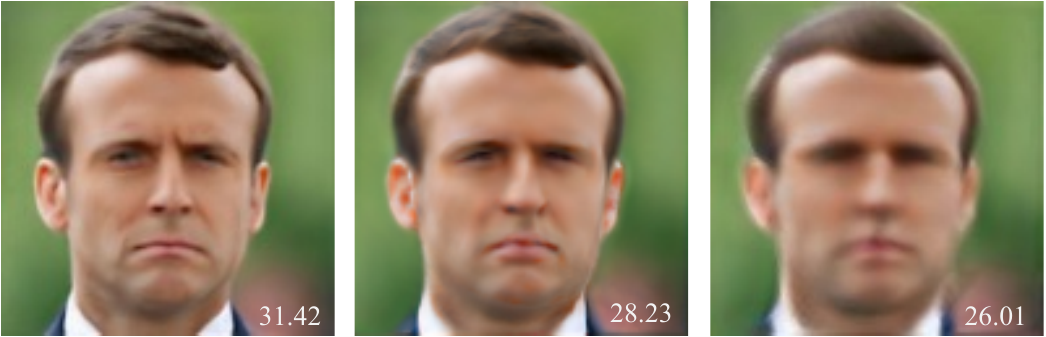} \vspace{-4pt}}
    \caption{{\bf Effect of Varying the Turbulence Strength $\mathbf{D/r_{0}}$.} We compare the the visual quality of a typical measurement (top row), the reconstruction by \cite{Mao2020Turbulence} (middle row), and by our method (bottom row) at 3 different $D/r_{0}$ levels from left to right. We present the PSNR at the bottom right of each image. TurbuGAN outperforms Mao et al.~\cite{Mao2020Turbulence} at all 3 turbulence strengths with sharper and more accurate reconstructions.} \label{fig:turbulence_strength}
\end{figure}

We now demonstrate TurbuGAN can be made robust to misspecified forward models by treating $D/r_{0}$ as a learnable parameter. 
To analyze the impact of learning to update $D/r_{0}$ in the simulator, we initialize the simulator with various $D/r_{0}$ values and then compare the final performance when one learns or fixes the $D/r_{0}$ parameter. In this set of experiments, we select the same face image as the target and set the true unknown $D/r_{0} = 2$ and test under five conditions with different initial $D/r_{0}$ ranging from 1 to 5. 

In Fig.~\ref{fig:learning_variances}, we plot the final reconstruction quality (measured in PSNR) against different misspecified initial $D/r_{0}$. 
Evidently, learning to adjust the turbulence strength level $D/r_{0}$ during training overcomes the incorrect initialization (we observed the adapted turbulence strengths consistently converged to the true $D/r_{0}$) and significantly improves the reconstruction quality.

\subsection{Results on Experimentally Captured Images}
For the experimentally-captured setting, we use images from the dataset provided by CVPR 2022 UG2 Challenge: Atmospheric Turbulence Mitigation~\cite{ug2challenge}.
These images are captured by the providers with turbulence generated in a real-world environment using heated hot air. \edit{The aperture diameter is 0.08125 meters; the target distance is 300 meters; and the Fried parameter, pixel-pitch, and focal-length are unknown.}

%\subsection{Face Images}
%\label{ssec:main_results}
Since the levels of turbulence strength are unknown for these data, we jointly learn to adjust our simulated turbulence strength level $D/r_{0}$ alongside the networks.
Reconstruction from these experimentally-captured data in Fig.~\ref{fig:realdata} demonstrates that TurbuGAN is able to recover images from real-world turbulence as well, though in the case of real turbulence the reconstructions are not appreciably better than~\cite{Mao2020Turbulence}. (No ground truth is provided.)

\begin{figure*}[t]
    \centering
    % \SevenColSubfig{./figures/plane/plane-895b_128.pdf}%
    \FiveColSubfig{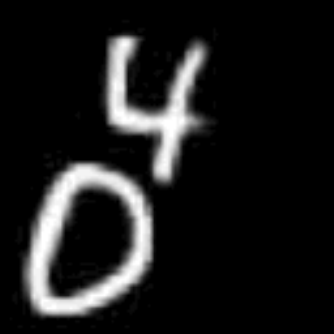}%
    \FiveColSubfig{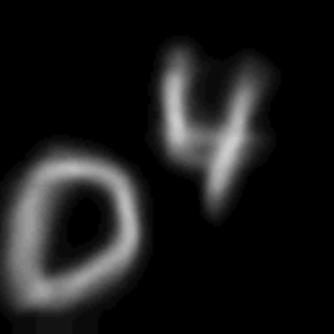}%
    \FiveColSubfig{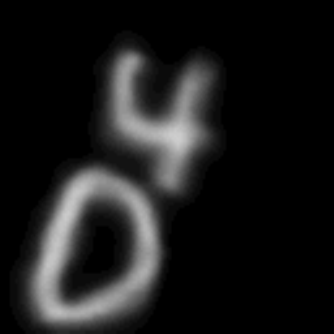}%
    \FiveColSubfig{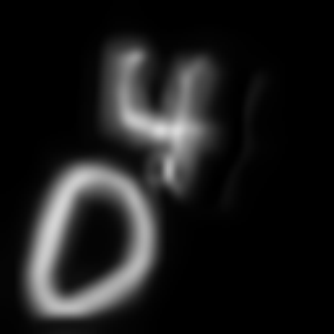}%
    \FiveColSubfig{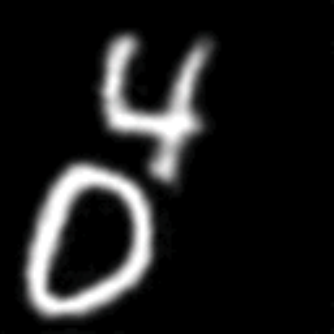}%

    \FiveColSubfigwCaption{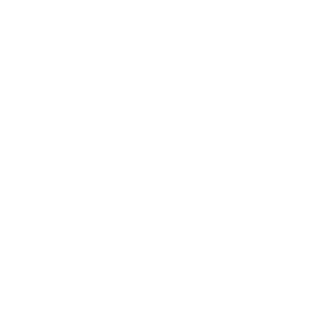}{Reference}%
    \FiveColSubfigwCaption{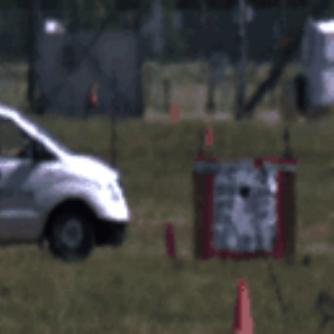}{Measurement}%
    \FiveColSubfigwCaption{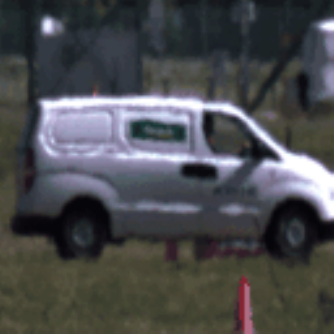}{Measurement}%
    \FiveColSubfigwCaption{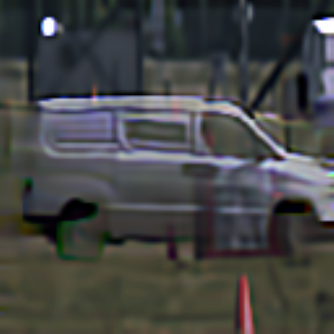}{Mao et al.}%
    \FiveColSubfigwCaption{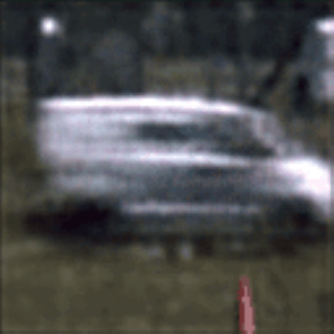}{Ours}%
    \caption{\edit{{\bf Dynamic Scene Reconstructions from Simulated (top) and Real-world (bottom) Turbulence.} Columns (b) and (c) show two measurements at different timestamps. Columns (d) and (e) show reconstructed frames at the same timestamp as (c).  For the simulated sequence with 256 frames, TurbuGAN produces sharper results than Mao et al.~\cite{Mao2020Turbulence}. For the real-world scene with 91 frames (acquired from~\cite{Clear2}), the reconstructions of both TurbuGAN and Mao et al.~\cite{Mao2020Turbulence} show visible artifacts, though Mao et al.'s reconstruction is considerably sharper.}} \label{fig:dynamic}	
	%\vspace{-5pt}
\end{figure*}

% \begin{figure*}[t]
%     \centering
%     % \SevenColSubfig{./figures/plane/plane-895b_128.pdf}%
%     \FiveColSubfig{./figures/rebuttal_van/measurement.png}%
%     \FiveColSubfig{./figures/airliner/lucky_airliner_hotair_18.pdf}%
%     \FiveColSubfig{./figures/airliner/stanley_airliner_hotair_18.pdf}%
%     \FiveColSubfig{./figures/airliner/XL_airliner_hotair_18.pdf}%
%     \FiveColSubfig{./figures/airliner/DIP_airliner_hotair_18.pdf}%

%     \FiveColSubfigwCaption{./figures/fighter/typical_fighter_hotair_41.pdf}{Typical}%
%     \FiveColSubfigwCaption{./figures/fighter/lucky_fighter_hotair_41.pdf}{Lucky}%
%     \FiveColSubfigwCaption{./figures/fighter/stanley_fighter_hotair_41.pdf}{Mao et al.}%
%     \FiveColSubfigwCaption{./figures/fighter/XL_fighter_hotair_41.pdf}{Ours-IN}%
%     \FiveColSubfigwCaption{./figures/fighter/DIP_fighter_hotair_41.pdf}{Ours-DIP}%
%     \caption{{\bf Reconstructions from Experimentally-Captured Turbulence-Distorted Images}. We further apply TurbuGAN to measurements distorted by real-world atmospheric turbulence. Each scene contains 100 measurements with unknown levels of turbulence distortion. Results show that TurbuGAN's performance is robust on real-world data. Our reconstructions when not assuming any domain-specific priors ({\it DIP}) appear sharper than when leveraging the prior from the ImageNet ({\it IN}) domain, as shown in (d) and (e).} \label{fig:realdata}
% 	%\vspace{-5pt}
% \end{figure*}

\subsection{Dynamic Scenes Reconstruction From Both Simulation and Real-World Measurements}\label{ssec:dynamic}
\edit{We also test our method on dynamic scenes with turbulence distortion. In this case, we modify the untrained U-Net generator so that it can take in the timestamp (replicated to match the image size) as an additional channel. Fig.~\ref{fig:dynamic} shows our results on 2 dynamic scenes. The top scene is computationally simulated by moving the digit 4 from the top left corner towards the bottom left corner, while fixing the digit 0 on the bottom left corner. We then simulate turbulence distortion~\cite{mao2021accelerating} on each frame under the turbulence strength $D/r0 = 2$. The bottom scene is a real-world turbulence-distorted video acquired from~\cite{moving_van}. Its target distance is 750 meters. All other parameters are unknown. While our method outperforms Mao et al.~\cite{Mao2020Turbulence} on simple simulated dynamics scenes, both methods struggle with real-world turbulence scenes. Video sequences of the measurements and reconstructions are included in the supplementary material.}%: our method is blurry, and Mao et al.~\cite{Mao2020Turbulence} suffers severe artifacts.} 

\begin{figure}[!th]
    \centering
    
   % \begin{overpic}[width=70pt]{./figures/zebra_sample/real_idx_12.png}
     % \put(7,33.5){$D/r_{0}=1$} 
     % \end{overpic}
   % \begin{subfigure}{70pt}
	%\centering
	%\begin{overpic}
	%\includegraphics[width=70pt, height=70pt]{./figures/zebra_sample/real_idx_12.png}
	%\put(7,33.5){$D/r_{0}=1$} 
    %\end{overpic}
    %\end{subfigure}
    \SixColSubfig{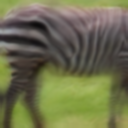}%
    ~
    \SixColSubfig{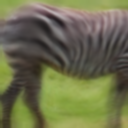}%
    ~
    \SixColSubfig{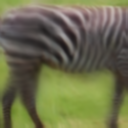}%
    ~
    \SixColSubfig{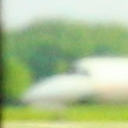}%
    ~
    \SixColSubfig{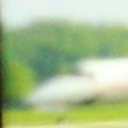}%
    ~
    \SixColSubfig{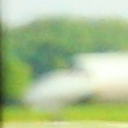}%
    \begin{subfigure}{\linewidth} \vspace{3pt}
    \caption{Observed Measurements} \vspace{-3pt}
    \end{subfigure}

    \SixColSubfig{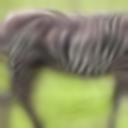}%
    ~
    \SixColSubfig{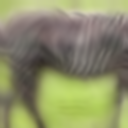}%
    ~
    \SixColSubfig{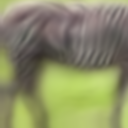}%
    ~
    \SixColSubfig{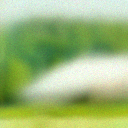}%
    ~
    \SixColSubfig{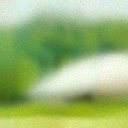}%
    ~
    \SixColSubfig{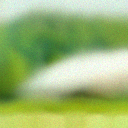}%

    \begin{subfigure}{\linewidth} \vspace{3pt}
    \caption{Simulated Measurements - Initial}\vspace{-3pt}
    \end{subfigure}

    \SixColSubfig{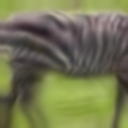}\hspace{-0mm} %
    ~
    \SixColSubfig{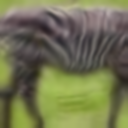}%
    ~
    \SixColSubfig{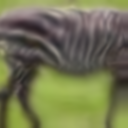}%
    ~
    \SixColSubfig{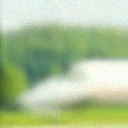}%
    ~
    \SixColSubfig{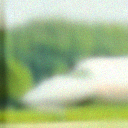}%ƒ
    ~
    \SixColSubfig{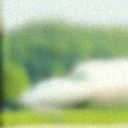}%
    \begin{subfigure}{\linewidth} \vspace{3pt}
    \caption{Simulated Measurements - Intermediate}\vspace{-3pt}
    \end{subfigure}

    \SixColSubfig{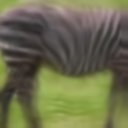}%
    ~
    \SixColSubfig{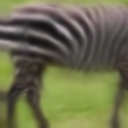}%
    ~
    \SixColSubfig{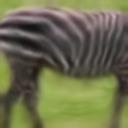}%
    ~
    \SixColSubfig{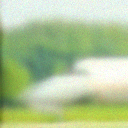}%
    ~
    \SixColSubfig{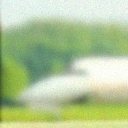}%
    ~
    \SixColSubfig{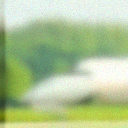}%
    \begin{subfigure}{\linewidth} \vspace{3pt}
    \caption{Simulated Measurements - Final}\vspace{-3pt}
    \end{subfigure}
\caption{{\bf Comparison Between the Distribution of Synthesized and Observed Measurements.} We show randomly selected frames from the observed measurements (a), simulated measurements at network initialization (b), simulated measurements at an intermediate iteration (c), and simulated measurements at the final iteration (d).
The distribution of the simulated measurements (b)-(d) becomes increasingly similar to the distribution of the observed measurements (a) over time. The zebra observations (left) are synthetic while the airplane observations (right) were experimentally captured. TurbuGAN gradually matches the distributions of both datasets.}
\label{fig:measurement_distribution}	%\vspace{-5pt}
\end{figure}

\begin{figure*}[!ht]
    \centering
    \includegraphics[width=0.48\linewidth]{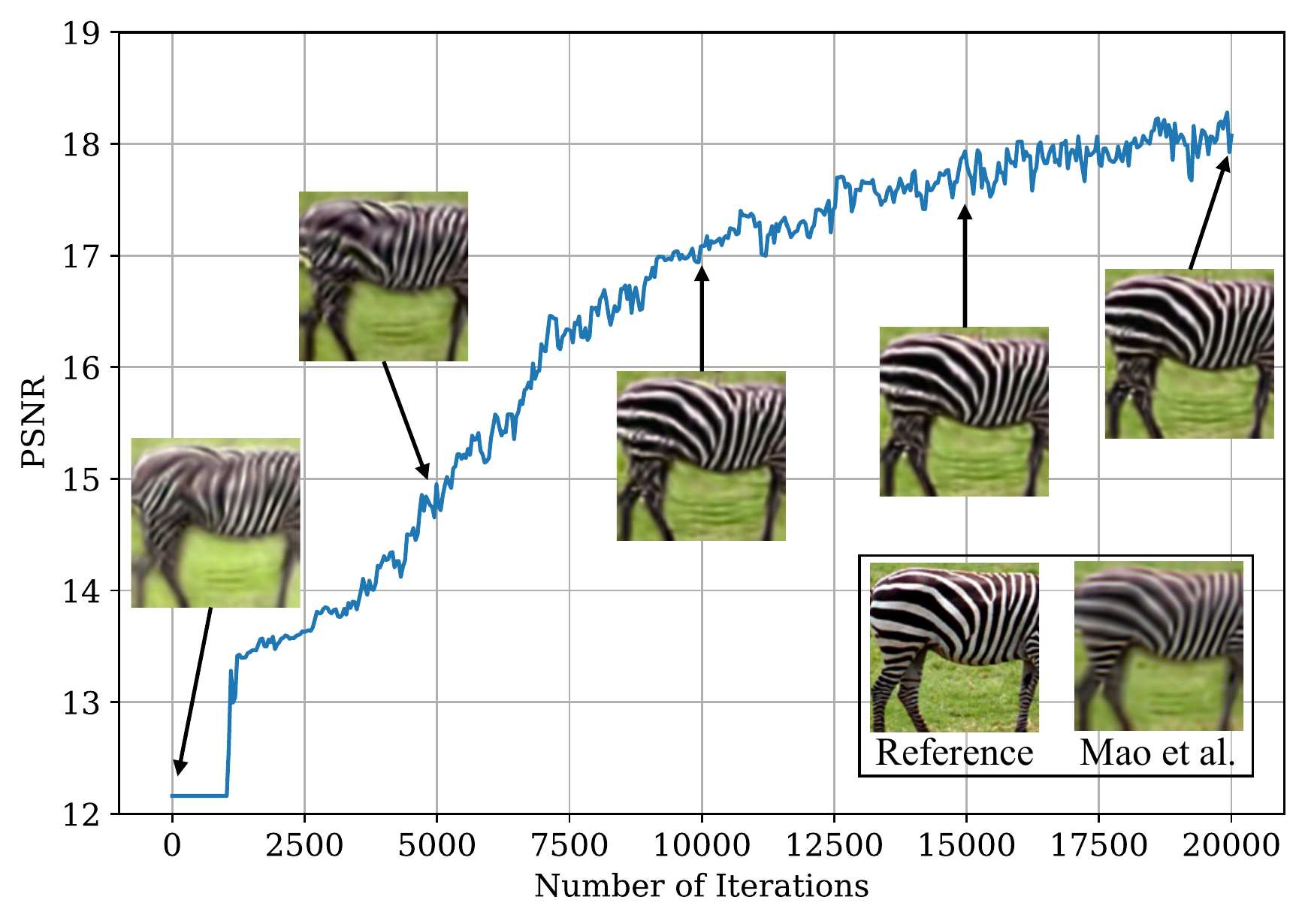}
    \includegraphics[width=0.48\linewidth]{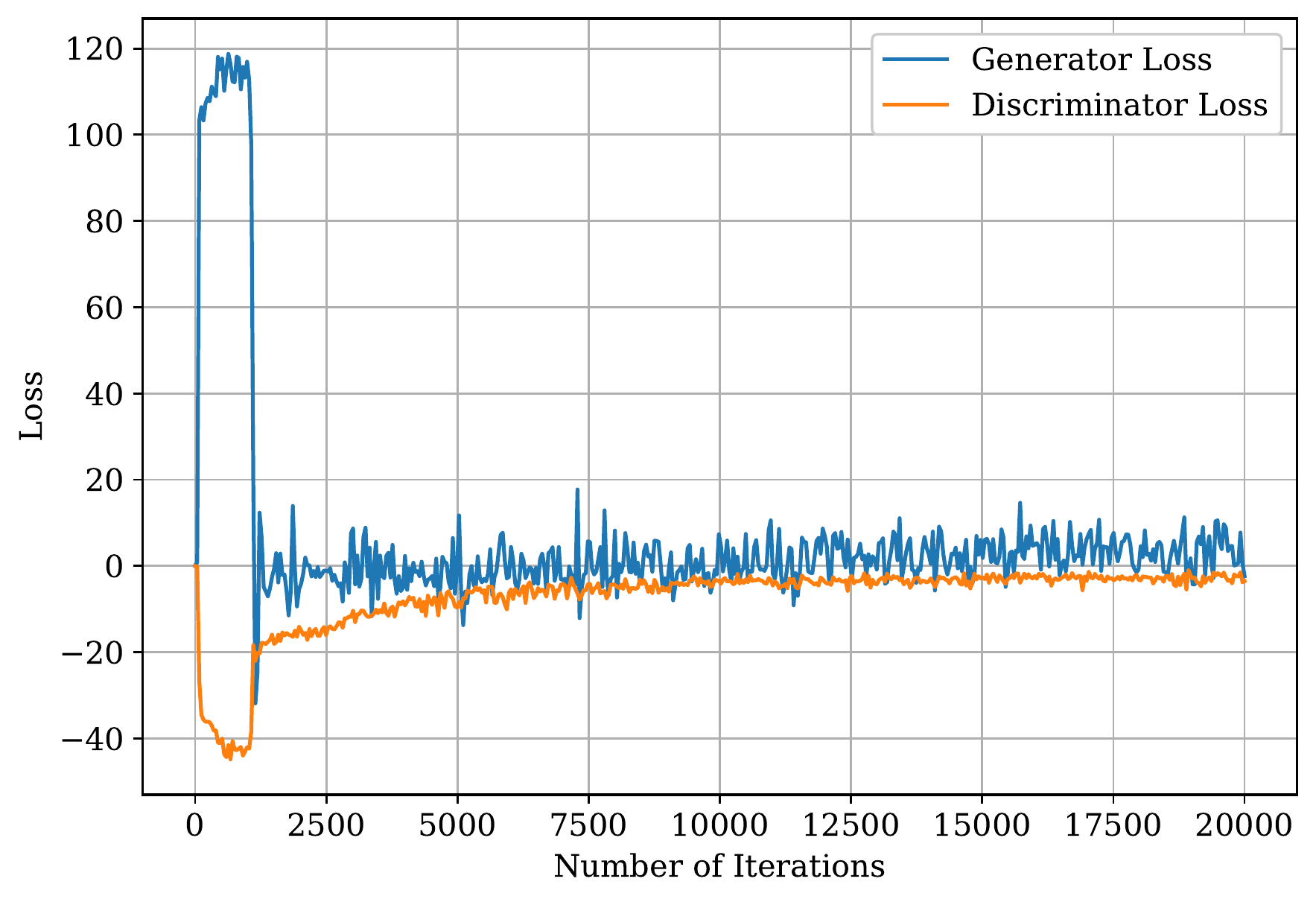}
    \caption{{\bf Reconstruction Improves During Training.} For the scene shown in Figure~\ref{fig:measurement_distribution}, we additionally plot the PSNR (left) and training losses (right). On the left, we also display the initial generator output and the reconstructions as training progresses. The reference ground truth image and the reconstruction by~\cite{Mao2020Turbulence} are placed at the lower right corner for comparison. Both the visual quality and PSNR steadily improve over time. On the right, we present the loss curves for both the generator and the discriminator, showing that the two networks gradually reach a stalemate where neither can improve much further.} \label{fig:convergence}
\end{figure*}
\section{Analysis and Discussion}
In this section, we present further analysis on how varying different attributes of the problem setting affect TurbuGAN's performance. Additionally, we discuss the implications of our method.

%\subsection{Ablation Analysis}
{\bf Different Turbulence Levels.}
We assess TurbuGAN on different levels of difficulty by varying the strength of turbulence when simulating the blurry measurements.
Fig.~\ref{fig:turbulence_strength} demonstrates that TurbuGAN consistently outperforms~\cite{Mao2020Turbulence} across a variety turbulence levels.

{\bf Distribution Matching.}
As noted in Section~\ref{ssec:theorem}, TurbuGAN's reconstruction capabilities hinge on its ability to successfully match the observed and simulated distributions. 
In Fig.~\ref{fig:measurement_distribution}, we demonstrate that during adversarial training, the distribution of the simulation measurements gradually becomes indistinguishable from the true measurements' distribution.
Fig.~\ref{fig:convergence} highlights how the reconstruction quality improves as TurbuGAN learns to improve the distribution of simulated measurements. 
We do not use early stopping since we consistently observe that as the number of iterations increases, the reconstruction quality steadily improves in terms of both visual quality and quantitative metrics such as PSNR.

\begin{figure}[!ht]
    \centering
    \SixColSubfigwCaption{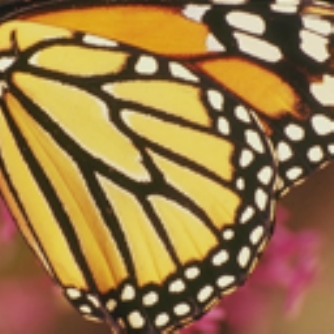}{Reference}%
    ~
    \SixColSubfigwCaption{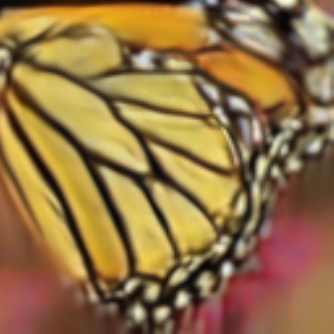}{$L=2$ $\vert$ 12.30}%
    ~
    \SixColSubfigwCaption{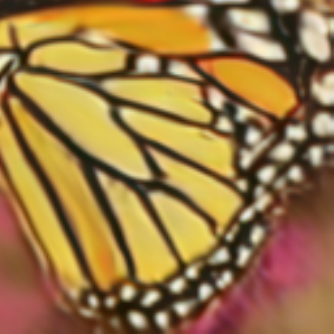}{$L=20$ $\vert$ 18.39}%
    ~
    \SixColSubfigwCaption{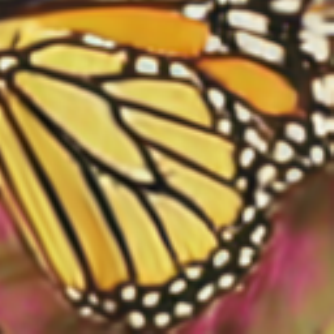}{$L=200$ $\vert$ 20.29}%
    ~
    \SixColSubfigwCaption{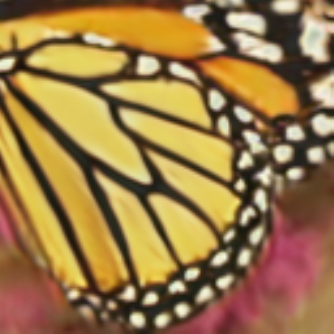}{$L=2000$ $\vert$ 21.46}%
    %~
    %\SixColSubfigwCaption{./figures/butterfly_1.5/stanley_N2000_19.18.pdf}{~\cite{Mao2020Turbulence} $\vert$ 19.18 dB}%
    
    \caption{{\bf Varying the Number of Measurements $L$.}
    We deploy TurbuGAN with a varying number of observations under simulated turbulence strength $D/r_{0}=1.5$. TurbuGAN benefits from more observations: 
    Both the visual quality and the PSNR of the reconstructions (displayed next to L) improve as $L$ increases.%, thus demonstrating TurbuGAN benefits from more observations.
    %(f) shows the result of Mao et al.~\cite{Mao2020Turbulence} using 2000 measurements. Note that when using 200 or more measurements, our results in (d) and (e) outperform (f) in terms of both PSNR and sharpness. 
    } \label{fig:num_measurements}	%\vspace{-5pt}
\end{figure}

\begin{figure}[!t]
    \centering
    \includegraphics[width=0.7\linewidth]{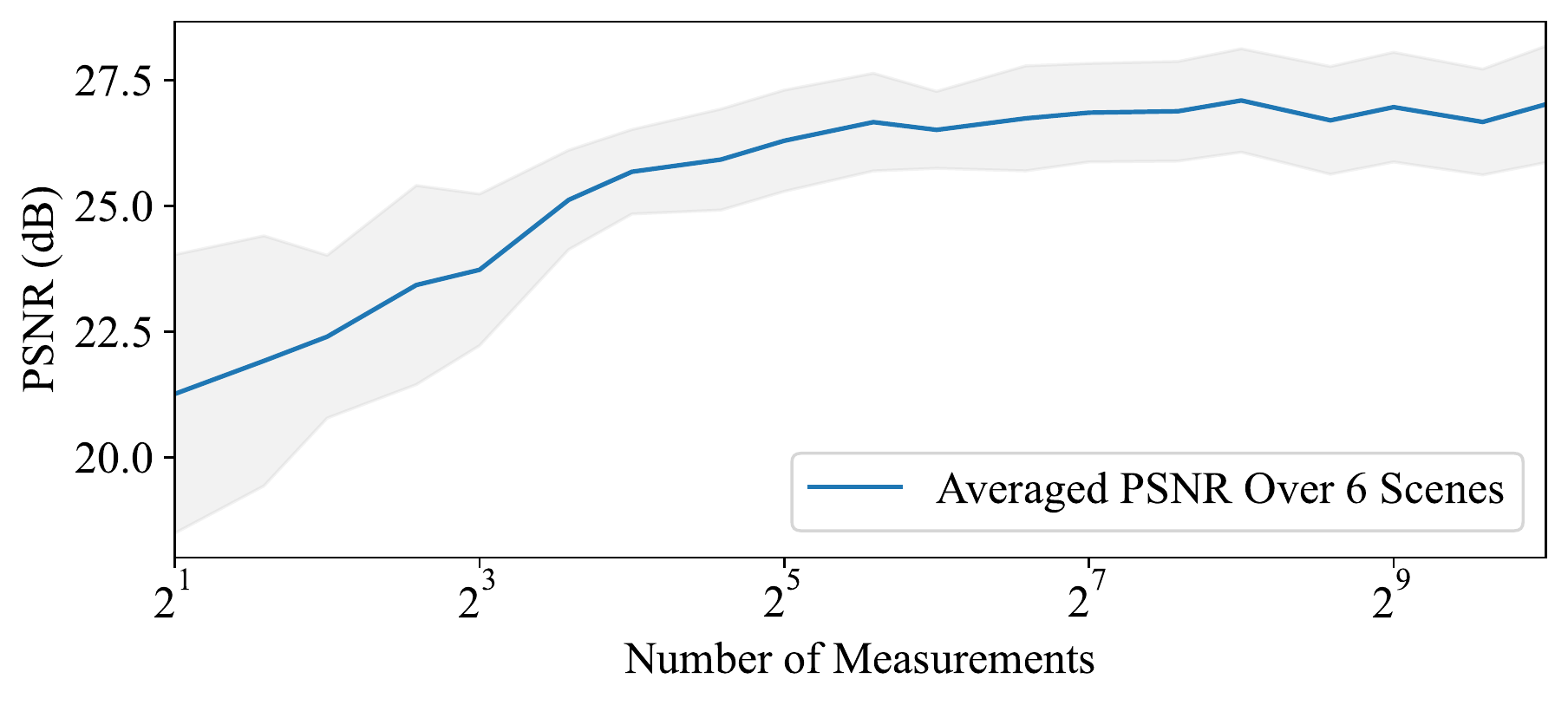}    
    \caption{\edit{{\bf Reconstructed PSNR Increases as the Number of Measurements Increases from 2 to 1024.} Six experiments were performed on different face targets, such as the ones in Fig.~\ref{fig:main_comparison}. The blue line illustrates the average PSNR, while the shaded gray area is all within one standard deviation of the mean. The PSNR improves gradually and exhibits no clearly defined phase transition.}}% This is a useful (negative) result which adds to the information theory literature and could prevent wasted effort: It suggests that, at least with the priors and algorithms used in our work, attempting to derive an RIP-like result for turbulence mitigation is a dead end.}} %
    \label{fig:psnr_no_phase_transition}
\end{figure}

{\bf Measurement Usage.}
Among learning-based methods, TurbuGAN is somewhat unique in its ability to take advantage of an arbitrary number of measurements. As demonstrated in Fig.~\ref{fig:num_measurements}, TurbuGAN's reconstruction quality increases with the number of available measurements. While our method can produce reasonable reconstructions using only 20 measurements, the more it has the better it does. \edit{ Moreover, as demonstrated in Fig.~\ref{fig:psnr_no_phase_transition}, when reconstruction is performed with adversarial sensing, the anisoplanatic turbulence removal problem does not demonstrate an obvious phase transition: Adding additional measurements tends to monotonically improve the reconstruction accuracy and there are no sharp increases in performance as one reaches a certain number of measurements.}

% \edit{
% \begin{finding}\label{find:mainone}
% \textbf{No Obvious Phase Transition.} As shown in Fig.~\ref{fig:psnr_no_phase_transition}, when reconstruction is performed with adversarial sensing, the anisoplanatic turbulence removal problem does not demonstrate an obvious phase transition: Adding additional measurements tends to monotonically improve the reconstruction accuracy and there are no sharp increases in performance as one reaches a certain number of measurements.\\
% \end{finding}
% }

% \edit{
% {\bf No Obvious Phase Transition. }As shown in Fig.~\ref{fig:psnr_no_phase_transition}, when reconstruction is performed with adversarial sensing, the anisoplanatic turbulence removal problem does not demonstrate an obvious phase transition: Adding additional measurements tends to monotonically improve the reconstruction accuracy and there are no sharp increases in performance as one reaches a certain number of measurements.\\
% }

%\subsection{Discussion}
{\bf Connections to EM.} Although the well-known EM algorithm can be also regarded as comparing the distributions of real and simulated measurements, in EM this comparison is performed explicitly using analytic expressions, rather than implicitly with a discriminator.
Accordingly, EM requires one to discretize the distribution of the latent variables (to make marginalization computationally feasible).
EM also requires analytically maximizing the log-likelihood of measurements, often resulting in a Gaussian assumption on the noise, which turns the log-likelihood maximization problem into an easy-to-solve least-squares problem.
Unfortunately, EM does not scale to high dimensional latent variables. 
For example, even with 15 Zernike coefficients each discretized into only 5 bins, each E step of the EM algorithm would require computing $5^{15}$ potential measurements per image! Modeling anisoplanatic turbulence is significantly harder still. 
TurbuGAN avoids the high computational complexity of the EM algorithm and is able to reconstruct the scene without explicitly solving for the continuous latent variables associated with each measurement.

% {\bf Assumptions on Blur Kernels.} 
{\bf Relation to Prior Theory.}
\edit{Prior work on multiframe/multichannel blind deconvolution problems has shown that it is possible to reconstruct a signal assuming that the blur kernels are sparse~\cite{Shan2008HighqualityMD, Levin2009UnderstandingAE, Bronstein2005BlindDO} or lie in a low-dimensional subspace~\cite{Tong1998MultichannelBI, Ahmed2014BlindDU}.
This work proves blind deconvolution is possible in a more general setting: We assume only that the blur kernels come from an arbitrary known distribution and have nonvanishing Fourier magnitudes in expectation.
%In contrast, this paper presents a method with a more general, and thus more challenging, set of assumptions than previous work. 
These requirements allow us to take advantage of decades of optics research that has given us accurate models for turbulence-induced blur kernels~\cite{Noll1976ZernikePA}.
%We assume the blur kernels come from a distribution based on the fact that kernel statistics and distributions are extensively studied and well-known for atmospheric turbulence~\cite{Noll1976ZernikePA}.
}

%\clearpage

\section{Conclusion}
This work introduces TurbuGAN, a spatially-varying multiframe blind deconvolution method based on the adversarial sensing concept~\cite{Gupta2021CryoGAN}.
TurbuGAN is able to effectively synthesize thousands of highly distorted images, each modeling the effects of severe atmospheric turbulence, into a single high-fidelity reconstruction of a scene. 
While TurbuGAN's performance is especially strong in domains where it has access to prior information (via a pretrained generative network), when combined with untrained networks, TurbuGAN is also surprisingly effective in domains where no such priors are available. Moreover, TurbuGAN can adapt to misspecified forward models.

Adversarial-sensing-based reconstruction methods, like TurbuGAN, represent a fundamentally new approach to solving inverse problems in imaging. They open up the possibility of applying deep learning to tackle previously unsolvable inverse problems in imaging where limited or no training data is available and only the distribution of the forward measurement model is known.

%\section*{Acknowledgments}
%This should be a simple paragraph before the References to thank those individuals and institutions who have supported your work on this article.
\bibliography{main.bib}
\bibliographystyle{IEEEtran}

\appendices
\section{Proof of Theorem~\ref{thm:1}}\label{sec:appendix}

The following provides a proof for Theorem~\ref{thm:1}, which states that by matching the distributions of the observed and simulated measurements, adversarial sensing will reconstruct the Fourier magnitude of $\pmb{x}$ accurately at all frequencies where the power spectral density of the blur kernels is non-zero.% that are not in the stop-band ($0$ frequency response) of all the blur kernels in the distribution of $p_h$.

\begin{proof}
If two random vectors have the same distribution, then their Fourier transforms will have the same distribution. Thus, if
\begin{align}
p_{\pmb{h}* \pmb{x}}(\pmb{y})=p_{\pmb{h}* \tilde{\pmb{x}}}(\pmb{y}),
\end{align}
then
\begin{align}\label{eqn:jointPDF}
p_{\pmb{H} \circ \pmb{X}}(\pmb{Y})=p_{\pmb{H}\circ \tilde{\pmb{X}}}(\pmb{Y}),
\end{align}
where $\pmb{H},~\pmb{X},~\tilde{\pmb{X}},$ and $\pmb{Y}$ denote the Fourier transforms of $\pmb{h},~\pmb{x},~\tilde{\pmb{x}},$ and $\pmb{y}$ respectively. (Recall that convolution in the spatial domain is equivalent to elementwise multiplication in the spatial frequency domain.)

\edit{
For arbitrary spatial frequencies $(K_u,K_v)$, let $X=\pmb{X}(K_u,K_v)$, $\tilde{X}=\tilde{\pmb{X}}(K_u,K_v)$, ${Y}={\pmb{Y}}(K_u,K_v)$, and  $H=\pmb{H}(K_u,K_v)$.
}

\edit{
Random variables with the same distributions have the same marginal distributions therefore Eq.~\eqref{eqn:jointPDF} implies
\begin{align}
p_{H X}({Y})=p_{H \tilde{X}}({Y}).
\end{align}
}

\edit{
Let $R=HX$ and $Q=H\tilde{X}$. 
Random variables with the same distributions have the same absolute moments, therefore 
\begin{align}
\mathbb{E}_R[|R|^2]=\mathbb{E}_Q[|Q|^2],
\end{align}
or equivalently
\begin{align}\label{eqn:ExpectationExplicit}
\int\displaylimits_{R=-\infty}^{\infty}|R|^2p_{HX}(R)dR=\int\displaylimits_{Q=-\infty}^{\infty}|Q|^2p_{H\tilde{X}}(Q)dQ.
\end{align}
}

\edit{
Both $X$ and $\tilde{X}$ are constants, therefore 
\begin{align}
p_{HX}(R)=
\begin{cases}
  p_{H}(R/X)  & \text{ if }X\neq 0,  \\
  \delta({R}) & \text{ if }X=0,
\end{cases}
\end{align}
and 
\begin{align}
p_{HX}(Q)=
\begin{cases}
  p_{H}(Q/\tilde{X})  & \text{ if }\tilde{X}\neq 0,  \\
  \delta({Q}) & \text{ if }\tilde{X}=0,
\end{cases}
\end{align}
where $\delta(\cdot)$ represents a Dirac delta function. 
}

%Assuming $p_H(\cdot)$ is not a Dirac ($H$ is not always 0), if $X$ is 0 then $\tilde{X}$ must also be 0 and if $\tilde{X}$ is 0 then ${X}$ must also be 0.

\edit{
\subsection*{Case 1: $X=0$ or $\tilde{X}=0$}
}

\edit{
If $X=0$ the left-hand side of Eq.~\eqref{eqn:ExpectationExplicit} becomes
\begin{align}
\int\displaylimits_{R=-\infty}^{\infty}|R|^2\delta(R)dR=0.
%\int\displaylimits_{(HX)=-\infty}^{\infty}|{H} {X}|^2p_{H}\biggl(\frac{(HX)}{X}\biggl)d(HX)=\int\displaylimits_{(H\tilde{X})=-\infty}^{\infty}|{H} \tilde{X}|^2p_{H}\biggl(\frac{(H\tilde{X})}{\tilde{X}}\biggl)d(H\tilde{X}),
\end{align}
If $p_H(\cdot)$ is not a Dirac centered at 0, the only way for the right-hand side of Eq.~\eqref{eqn:ExpectationExplicit} to be $0$ is for $\tilde{X}$ to also be 0. One may similarly show $\tilde{X}=0$ implies $X=0$. 
}

\edit{
The assumption that $p_H(\cdot)$ is not a Dirac centered at 0 holds so long as the power spectral density of $\pmb{h}$ is non-zero at $(K_u,K_v)$: $\mathbb{E}_{H}[|H|^2]\neq 0 $ implies $p_H(\cdot)$ is not a Dirac centered at $0$. Thus, for frequencies $(K_u,K_v)$ where the power spectral density of $\pmb{h}$ is non-zero, we will recover $X$ and $\tilde{X}$ (and thus also their magnitudes) if either is $0$.
}
%As this proof is only concerned with frequencies where the power-spectral density of $\pmb{h}$ is non-zero the $p_H(\cdot)$; $\mathbb{E}_{H}[|H|^2]\neq 0 $ implies $p_H(\cdot)$ is not a Dirac.

%$p_{HX}(R)=p_{H}(R/X)$ and  $p_{H\tilde{X}}(Q)=p_{H}(Q/\tilde{X})$ and

\edit{
\subsection*{Case 2: Neither $X$ nor $\tilde{X}$ are $0$}
Alternatively, if neither $X$ nor $\tilde{X}$ are $0$ we may rewrite Eq.~\eqref{eqn:ExpectationExplicit} as
\begin{align}
\int\displaylimits_{R=-\infty}^{\infty}|R|^2p_{H}(R/X)dR=\int\displaylimits_{Q=-\infty}^{\infty}|Q|^2p_{H}(Q/\tilde{X})dQ.
%\int\displaylimits_{(HX)=-\infty}^{\infty}|{H} {X}|^2p_{H}\biggl(\frac{(HX)}{X}\biggl)d(HX)=\int\displaylimits_{(H\tilde{X})=-\infty}^{\infty}|{H} \tilde{X}|^2p_{H}\biggl(\frac{(H\tilde{X})}{\tilde{X}}\biggl)d(H\tilde{X}),
\end{align}
%using the fact that $p_{HX}(Y)=p_{H}(Y/X)$ and  $p_{H\tilde{X}}(Y)=p_{H}(Y/\tilde{X})$.
%We may ignored the corner cases $X=0$ and $\tilde{X}=0$ because
}

\edit{
Substituting $HX$ and $H\tilde{X}$ for $R$ and $Q$  we are left with
\begin{align}
\text{sign}(X)\int\displaylimits_{H=-\infty}^{\infty}|{H} {X}|^2p_{H}(H)XdH=\text{sign}(\tilde{X})\int\displaylimits_{H=-\infty}^{\infty}|{H} \tilde{X}|^2p_{H}(H)\tilde{X}dH,
\end{align}
where the leading sign terms ensure the proper limits of integration.
}

\edit{
This implies
\begin{align}
|X|^3\int\displaylimits_{H=-\infty}^{\infty}|{H}|^2p_{H}(H)dH=|\tilde{X}|^3\int\displaylimits_{H=-\infty}^{\infty}|{H}|^2p_{H}(H)dH,
\end{align}
or equivalently
\begin{align}
{|X|}^3\mathbb{E}[|{H}|^2]=\tilde{|X|}^3\mathbb{E}[|{H}|^2].
\end{align}
}

\edit{
This relationship holds for all spatial frequencies, which implies
% \newpage
% Random variables with the same distributions have the same absolute moments, therefore we have that
% \begin{align}
% \mathbb{E}[|\pmb{H}\circ \pmb{X}|^2]=\mathbb{E}[|\pmb{H} \circ \tilde{\pmb{X}}|^2].
% \end{align}
% This implies
% \begin{align}\label{eqn:PSD_relation}
% |\pmb{X}|^3\circ\mathbb{E}[|\pmb{H}|^2]=|\tilde{\pmb{X}}|^3\circ\mathbb{E}[|\pmb{H}|^2].
% \end{align}
% Equation~\eqref{eqn:PSD_relation} implies
\begin{align}\label{eqn:condition}
    |\pmb{X}|^3(K_u,K_v) &= |\tilde{\pmb{X}}|^3(K_u,K_v)~\forall(K_u,K_v) \text{ s.t. }\mathbb{E}[|\pmb{H}|^2](K_u,K_v)\neq 0.
\end{align}
}

\edit{
Let $\pmb{S}$ denote a frequency domain mask with entries equal to $0$ at spatial frequencies $(K_u,K_v)$ where the power spectral density of $\pmb{h}$ is $0$, and with entries equal to $1$ everywhere else. That is, 
\begin{align}
    \pmb{S}(K_u,K_v)=\begin{cases}
    1,\text{ if } \mathbb{E}[|\pmb{H}|^2](K_u,K_v)>0,
    \\
    0,\text{ otherwise. } 
    \end{cases}
\end{align}
We can now concisely represent condition Eq.~\eqref{eqn:condition} with the expression
\begin{align}
    \pmb{S}\circ|\pmb{X}|^3=\pmb{S}\circ|\tilde{\pmb{X}}|^3,
\end{align}
which implies
\begin{align}
    \pmb{S}\circ|\pmb{X}|=\pmb{S}\circ|\tilde{\pmb{X}}|.
\end{align}
}
\end{proof}

%\appendices
\section*{Proof of Corollary~\ref{corr:1}}\label{sec:appendix2}
The following provides a proof for Corollary~\ref{corr:1}, which states that if, in addition to assuming $p_{\pmb{h}* \pmb{x}}(\pmb{y})=p_{\pmb{h}* \tilde{\pmb{x}}}(\pmb{y})$, one assumes $\mathbb{E}[\mathcal{F}\pmb{h}](K_u,K_v)\neq 0$ for all spatial frequencies $(K_u,K_v)$, then one will successfully reconstruct $\pmb{x}$.

\begin{proof}
If two random vectors have the same distribution, then their Fourier transforms will have the same distribution. Thus, if
\begin{align}
p_{h* x}(\pmb{y})=p_{h* \tilde{x}}(\pmb{y}),
\end{align}
then
\begin{align}
p_{H \circ X}(\pmb{Y})=p_{H\circ \tilde{X}}(\pmb{Y}),
\end{align}
where $\pmb{H},~\pmb{X},~\tilde{\pmb{X}},$ and $\pmb{Y}$ denote the Fourier transforms of $\pmb{h},~\pmb{x},~\tilde{\pmb{x}},$ and $\pmb{y}$ respectively.

\edit{
Let the scalars $X$, $\tilde{X}$, ${Y}$, $\tilde{Y}$,  $H$, $R$, and $Q$ be defined as in the previous section.
}

\edit{
Random variables with the same distributions have the same moments, therefore we have that
\begin{align}
\mathbb{E}[R]=\mathbb{E}[Q],
\end{align}
which is equivalent to
\begin{align}\label{eqn:ExpectationExplicit2}
\int\displaylimits_{R=-\infty}^{\infty}Rp_{HX}(R)dR=\int\displaylimits_{Q=-\infty}^{\infty}Qp_{H\tilde{X}}(Q)dQ.
\end{align}
}

\edit{
If either $X$ or $\tilde{X}$ are equal to $0$ we can follow the steps of Case 1 in the previous proof to show they are equivalent. (The strong assumption, $\mathbb{E}[\mathcal{F}\pmb{h}](K_u,K_v)\neq 0$, used in this proof ensures the weaker assumption, $\mathbb{E}[|\mathcal{F}\pmb{h}|^2](K_u,K_v)\neq 0$, required for Case 1 of the previous proof holds.)
}

\edit{
Otherwise, we may rewrite Eq.~\eqref{eqn:ExpectationExplicit2} as
\begin{align}
\int\displaylimits_{R=-\infty}^{\infty}Rp_{H}(R/X)dR=\int\displaylimits_{Q=-\infty}^{\infty}Qp_{H}(Q/\tilde{X})dQ.
%\int\displaylimits_{(HX)=-\infty}^{\infty}|{H} {X}|^2p_{H}\biggl(\frac{(HX)}{X}\biggl)d(HX)=\int\displaylimits_{(H\tilde{X})=-\infty}^{\infty}|{H} \tilde{X}|^2p_{H}\biggl(\frac{(H\tilde{X})}{\tilde{X}}\biggl)d(H\tilde{X}),
\end{align}
}

\edit{
Substituting $HX$ and $H\tilde{X}$ for $R$ and $Q$  we are left with
\begin{align}
\text{sign}(X)\int\displaylimits_{H=-\infty}^{\infty}H {X}p_{H}(H)XdH=\text{sign}(\tilde{X})\int\displaylimits_{H=-\infty}^{\infty}H \tilde{X}p_{H}(H)\tilde{X}dH,
\end{align}
where the leading sign terms ensure the proper limits of integration.
}

\edit{
This expression simplifies to 
\begin{align}
{X\cdot|X|\cdot}\mathbb{E}[{H}]=\tilde{X}\cdot\tilde{|X|}\cdot\mathbb{E}[{H}],
\end{align}
which implies
\begin{align}\label{eqn:EhX}
\mathbb{E}[\pmb{H}]\circ \pmb{X}=\mathbb{E}[\pmb{H}] \circ \tilde{\pmb{X}}. 
\end{align}
}

Under the assumption $\mathbb{E}[\pmb{H}](K_u,K_v)\neq 0$ for all spatial frequencies $(K_u,K_v)$, Eq.~\eqref{eqn:EhX} implies 
$\pmb{X}=\tilde{\pmb{X}}$, which implies $\pmb{x}=\tilde{\pmb{x}}$.
\end{proof}

\section*{Proof of Corollary~\ref{corr:2}}\label{sec:appendix3}
The following provides a proof for Corollary~\ref{corr:2}, which considers the case where one misspecifies the forward model and ensures that $p_{\pmb{h}* \pmb{x}}(\pmb{y})=p_{\pmb{h}_{s}* \tilde{\pmb{x}}}(\pmb{y})$. The corollary states that if one assumes that $\pmb{h}_s=\pmb{h}_{correction}*\pmb{h}$ for all $\pmb{h}\sim p_{\pmb{h}}$ and that $\mathbb{E}[\mathcal{F}\pmb{h}_{correction}](K_u,K_v)\neq 0$ and $\mathbb{E}[\mathcal{F}\pmb{h}](K_u,K_v)\neq 0$ for all spatial frequencies $(K_u,K_v)$, then the reconstruction will be related to the true signal through $\tilde{\pmb{x}}=\pmb{h}_{correction}*\pmb{x}$.

 \begin{proof}
 The condition $p_{\pmb{h}* \pmb{x}}(\pmb{y})=p_{\pmb{h}_{s}* \tilde{\pmb{x}}}(\pmb{y})$
 implies
 \begin{align}
 p_{\pmb{h}_s* \pmb{h}_{correction}*\pmb{x}}(\pmb{y})=p_{\pmb{h}_{s}* \tilde{\pmb{x}}}(\pmb{y}).
 \end{align}

Similarly, $\mathbb{E}[\mathcal{F}\pmb{h}_{correction}](K_u,K_v)\neq 0$ and $\mathbb{E}[\mathcal{F}\pmb{h}](K_u,K_v)\neq 0$ implies 
\begin{align}
\mathbb{E}[\mathcal{F}\pmb{h}_s](K_u,K_v)\neq 0.
\end{align}

From here we can rely upon on the proof of Corollary~\ref{corr:1}, with $\pmb{h}$ replaced with $\pmb{h}_s$, to show that 
\begin{align}
\tilde{\pmb{x}}=\pmb{h}_{correction}*\pmb{x}.
\end{align}

 \end{proof}

\section{PSNR Values for Fig. 2}
\begin{table}[!ht]
\centering
\begin{tabular}{c c c c c c}
\toprule
Typical & Lucky & Mao et al. & Ours-Face & Ours-IN & Ours-DIP \\ \midrule
19.05 & 16.90  &   22.07  &  {\bf 30.54}  &   24.09  &   27.10   \\
22.10 & 19.60  &   24.80  &  {\bf 27.77}  &   24.31  &   24.98   \\
21.06 & 17.28  &   23.40  &  {\bf 26.87}  &   22.17  &   24.83   \\
23.04 & 19.88  &   26.61  &  {\bf 30.96}  &   26.83  &  29.92   \\
\bottomrule
\end{tabular}
\caption{Rows ordered as in Fig. 2.}
\end{table}

\section{PSNR Values for Fig. 3}
\begin{table}[!ht]
\centering

\begin{tabular}{c c c c c c}
\toprule
Typical & Lucky & Mao et al. & Ours-Face & Ours-IN & Ours-DIP \\ \midrule
22.80 & 20.86  &   25.10  &  24.27  &   24.58  &   {\bf 25.73}   \\
18.33 & 13.96  &   21.57  &  {\bf 24.91}  &   20.67  &   24.88   \\
11.76 & 9.96  &   13.08  &  14.88  &   13.64  &   {\bf 22.43}   \\
14.15 & 12.10  &   17.12  &  15.30  &   {\bf 19.81}  &  19.66   \\
16.67 & 13.99  &   19.29  &  23.65  &   19.15  &  {\bf 26.82}   \\
\bottomrule
\end{tabular}
\caption{Rows ordered as in Fig. 3.}
\end{table}

\end{document}